\documentclass{article}

\usepackage{amsmath}
\usepackage{amssymb}

\usepackage{arxiv}

\usepackage[utf8]{inputenc} 
\usepackage[T1]{fontenc}    
\usepackage{hyperref}       
\usepackage{url}            
\usepackage{booktabs}       
\usepackage{amsfonts}       
\usepackage{nicefrac}       
\usepackage{microtype}      
\usepackage{cleveref}       
\usepackage{lipsum}         
\usepackage{graphicx}
\usepackage{natbib}
\usepackage{doi}

\usepackage[ruled,vlined]{algorithm2e}

\title{Approximate Bayesian Computation Based on Maxima Weighted Isolation Kernel Mapping}

\author{ \href{https://orcid.org/0000-0002-7935-6776}{\includegraphics[scale=0.06]{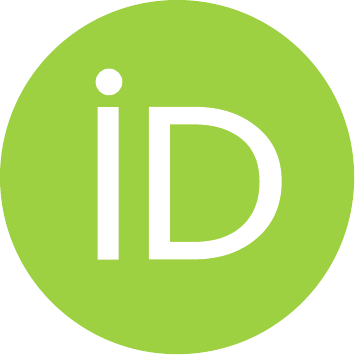}\hspace{1mm}Iurii S.~Nagornov}
\thanks{adress: 2-4-7 Aomi, Koto-ku, Tokyo 135-0064, Japan} \\
	NEC-AIST AI Cooperative Research Laboratory, \\
	Artificial Intelligence Research Center, \\
	The National Institute of Advanced Industrial Science and Technology \\
	 \texttt{iurii.nagornov@aist.go.jp} 
	}

\renewcommand{\shorttitle}{ABC based on Maxima Weighted Isolation Kernel Mapping}

\begin{document}
\maketitle

\abstract{\textbf{Motivation:} 
A branching processes model yields an unevenly stochastically distributed dataset that consists of sparse and dense regions.
This work addresses the problem of precisely evaluating parameters for such a model.
Applying a branching processes model to an area such as cancer cell evolution faces a number of obstacles, including high dimensionality and the rare appearance of a result of interest. We take on the ambitious task of obtaining the coefficients of a model that reflects the relationship of driver gene mutations and cancer hallmarks on the basis of personal data regarding variant allele frequencies. \\
\textbf{Results:} An approximate Bayesian computation method based on Isolation Kernel is developed. The method involves the transformation of row data to a Hilbert space (mapping) and the measurement of the similarity between simulated points 
and maxima weighted Isolation Kernel mapping related to the observation point. We also design a heuristic algorithm for parameter estimation that requires no calculation and is dimension independent. The advantages of the proposed machine learning method are illustrated using multidimensional test data as well as a specific example focused on cancer cell evolution.  
}

\keywords{Bayesian inference \and  Isolation Kernel \and Maxima Weighted Mapping \and Cancer cell evolution}


\section{Introduction}

{\bf Problem definition and motivation.}
This paper addresses the problem of precisely estimating the parameters of a stochastic model corresponding to branching processes.
A branching process is a stochastic process consisting of collections of random variables indexed by the natural numbers. 
Branching processes are often used to describe population models \cite{Branching1989} and \cite{Branching2012Review}; 
for example, models in population genetics showing genetic drift \cite{GeneticDrift2016} \cite{GeneticDrift2017}.  
In contrast to statistical approaches, branching processes enable the study of the dynamics of cell evolution and, as a consequence, have become a popular approach to cancer cell evolution research \citealp{SIAMReview2016}. 
However, particularly in the case of cancer cell evolution, as well as in branching processes in general, the ultimate extinction of a population often occurs \cite{Branching1998}. 
It is for this reason that depending on the initial uniform distribution of parameters, branching processes models tend to yield unevenly distributed data consisting of sparse and dense regions.

The stochastic nature of the data is an another obstacle in estimating the parameters of a branching processes model, especially in the case of cancer cell evolution \cite{Dataset_tugHall2021}. 
Moreover, simulations, based on a model of cell mutations, population evolution, and tumor/cancer subpopulations, commonly lead to the emergence of many clones and rarely to the appearance of cancer cells.
The ambitious task of obtaining accurate coefficients in a model reflecting the relationship of driver genes mutations and cancer hallmarks on the basis of personal data of variant allele frequencies (VAFs) is defined in \cite{tugHall_bioinf2020}. 
Solving this challenging problem with precise parameters offers the possibility of making personal predictions regarding the development of cancer when a certain mutated gene is blocked by a particular drug.
To obtain model parameters that reflect the personal characteristics of cancer evolution, it is necessary to develop a method for their assessment based on rare events and extremely unevenly distributed stochastic simulation data.

{\bf Proposal of a decision.}
Approximate Bayesian Computation (ABC) is a well-known approach to estimating model parameters under a given observation \cite{ABCPrictice2010}. 
For example, in \citealp{subclonal2018}, where the authors quantify subclonal selection, ABC was applied to an observation dataset consisting of VAFs at the genome level  in multiple cancers. 
Unfortunately, it has been found that rejection ABC produces inaccurate estimates in the case of unevenly distributed data in a multidimensional space of parameters \cite{KernelABC2013}. 
The precise estimation of parameter values lying in the sparse region is problematic when the data are widely scattered. 
In light of this limitation, we propose the use of a machine learning method, combined with ABC, to improve estimation accuracy.
Because of its unique properties, Isolation Kernel (iKernel) would appear to be the most appropriate method for this class of problem \citealp{iKernel_dist2020}.

Isolation Kernel was first proposed in \citealp{iKernel_SVM2018}, where the support vector machine problem was solved in the multidimensional case more successfully than with other methods. 
The advantages of iKernel are its low computational cost \citealp{iKernel_dist2020}, the possibility of evading the curse of dimensionality \citealp{BreakingCurseD2021} and, most interestingly, its data dependence.  
Meaning that the similarity between two data points depends on the density of the dataset around these points (here and hereafter, by similarity we mean the similarity calculated using iKernel if another method is not indicated).

\begin{figure}[!tpb]
\centerline{\includegraphics[width=85mm]{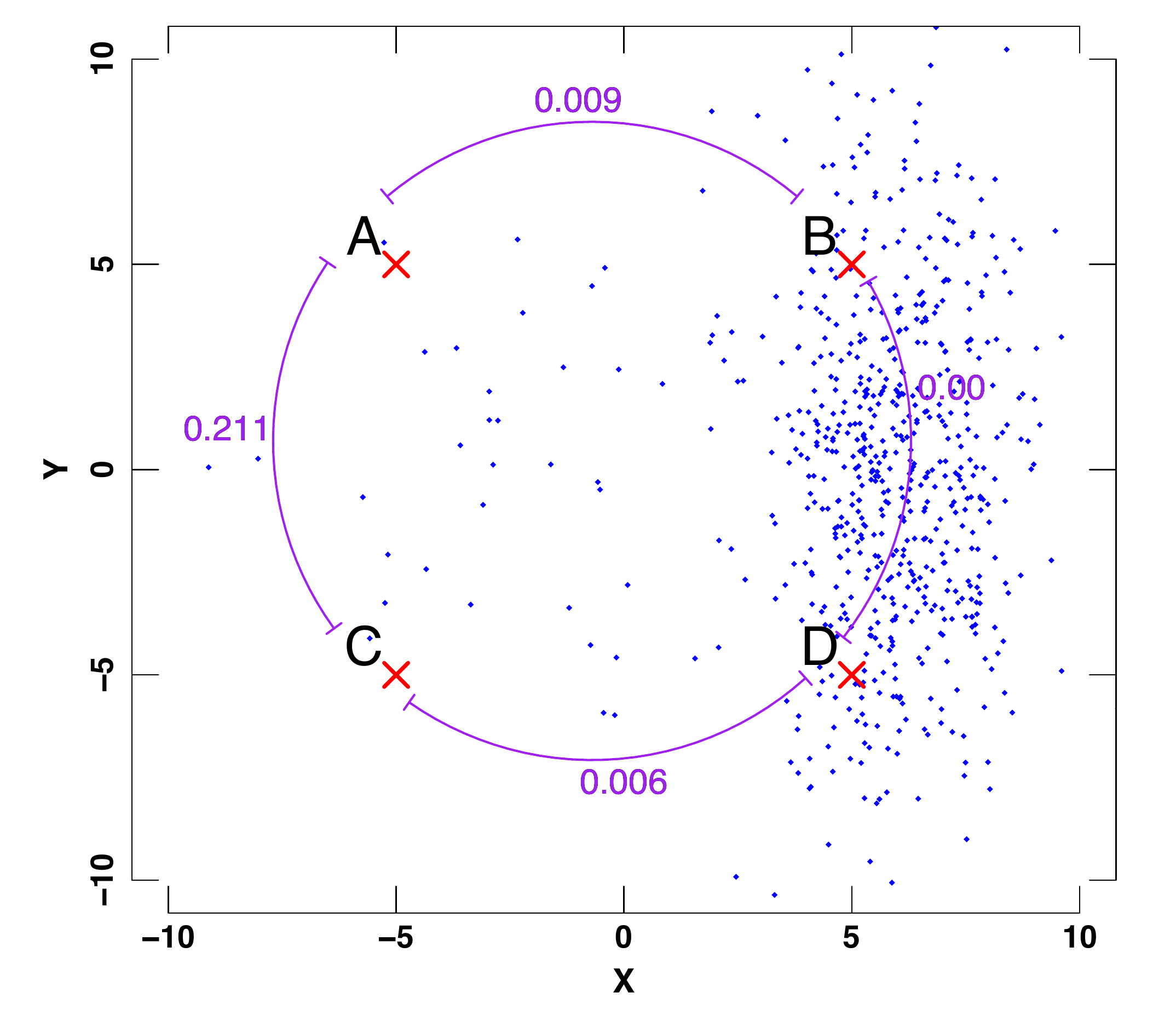}}
\caption{Demonstration of the dependence of Isolation Kernel on density in a two-dimensional dataset represented by points.
Crosses A,B,C, and D are equally distant points: $d_{AB} = d_{BD} = d_{CD} = d_{AC}$ 
The right region is dense, the left is sparse.
Arcs with numbers show the similarities between points.}
\label{fig:01}
\end{figure}

To demonstrate this last property, consider unevenly distributed data with sparse and dense regions (Fig.~\ref{fig:01}).
Let us select four points equally distant within these regions: A,B,C, and D.
Ignoring for now the specific algorithm for calculating similarity $s \in [0,1]$, we can simply show the calculation of $s$ between the different points: 
for B and D, $s_{BD} = 0$ in the dense area;
in contrast, $s_{AC}$ is relatively large, with a value of 0.211 ($s = 1$ indicates identical points). 
The similarity between points from the different areas is very small but not zero  ($s_{AB} = 9\cdot 10^{-3}$ and $s_{CD} = 6\cdot 10^{-3}$).
Thus, similarity here includes the inverse distance as well as the number of neighbor points, a property used in ABC based on Isolation Kernel mapping. 

It is important to consider the principles of feature mapping in discussing the proposed method. 
For this reason, we describe kernel mapping in the next section.
The remainder of the paper proceeds as follows:
Section 2 provides the background for our research, with a brief overview of kernel methods and the specific properties of feature mapping and kernel ABC.
Section 3 comprises details of the proposed ABC method based on Isolation Kernel and the Voronoi diagram; the maxima weighted iKernel mapping is defined and a heuristic algorithm for parameter estimation is presented.
Section 4 illustrates the proposed method using multidimensional synthetic examples and an application of the method to parameter estimation for a simulator of cancer cell evolution.
Section 5 offers concluding remarks.


\section{Background}

\subsection{Nonlinear feature map}

Kernel functions have become increasingly popular in machine learning (ML) due to improvements in statistical inference methods \citealp{KernelTest2012} and \citealp{KernelConditional2008}.
The ML tasks reach success under critical condition on the accuracy of the estimation of the data-generating process, and kernel methods can be used to significantly improve the accuracy and reproducibility of this process \citealp{ReviewKernel2017}. 
The general idea of a kernel method is to transform real data $x \in \mathcal{X}$ to a Hilbert space in which a similarity measure is used instead of a distance.
The similarity measure is usually non-linear and much more flexible \citealp{KernelBayesRule2013}; for example, similarity, in contrast to distance metrics, does not obey the triangle inequality rule. 

As part of the kernel method, Hilbert space $\mathcal{H}$ is a vector space with an inner product $\langle \cdot, \cdot \rangle$ and feature transformation from real data $\phi(x): \mathcal{X} \rightarrow \mathcal{H}$ also called mapping \citealp{ReviewKernel2017}. 
A Hilbert space typically has a dimensionality $d_{\mathcal{H}}$ that is much higher than the dimension $d$ of the original real data space $\mathcal{X}$.

The inner product plays a role in the similarity function in $\mathcal{H}$
and is represented by a kernel function such that $k(x,y) = \langle \phi(x), \phi(y) \rangle$ for any points $x,y \in R^d$ and $\phi(x), \phi(y) \in \mathcal{H}$ with dimension $d_\mathcal{H}$ \citealp{ReviewKernel2017}.
For ML, the kernel is chosen as a positive definite function with symmetry, i.e. $k(x, y) = k(y, x)$ \citealp{KernelBayesRule2013}. 
Positive definiteness is a necessary condition for choosing the kernel. 
It is important to note that all kernels based on distances are positive definite and symmetric. 

The feature mapping here includes data transformation with increasing dimensionality $d_{\mathcal{H}} \gg d$. In principle, dimension of a Hilbert space can be infinite, i.e.,
$d_{\mathcal{H}} = \infty$, as, for example, in the case of Laplacian and Gaussian kernels. 
It is for this reason that the explicit calculation of $\phi(x)$ is typically a difficult or impossible computational task. 
The most attractive feature of kernel methods is that they avoid the explicit calculation of $\phi(x)$ and instead use only the kernel function. This procedure is commonly called a kernel trick and is a very convenient, efficient and accurate algorithm \citealp{ReviewKernel2017}. 
Thus, one can calculate the similarity between all data points in the form of a Gram matrix that is also positive definite, defined as follow:

\begin{equation}
G =
\begin{pmatrix}
k(x_1, x_1) & \cdots & k(x_1, x_n) \\
\vdots &  \ddots  & \vdots \\
k(x_n, x_1) & \cdots & k(x_n, x_n) 
\label{eq:01}
\end{pmatrix}
\end{equation}

For ML, a reproducing kernel Hilbert space (RKHS) with a reproducing kernel having the following reproducing property is used:
\vspace*{-4pt}
\begin{equation}
f (x) = \langle f ,k(\cdot,x) \rangle
\label{eq:02}
\end{equation}
for any function $f(x) \in \mathcal{H}$ and any point $x \in R^d$ where the dot in $k(\cdot,x)$ indicates all possible points in a Hilbert space. There is a one-to-one correspondence between the reproducing kernel $k$ and the RKHS $\mathcal{H}$ (Theorem 2.5 in \citealp{ReviewKernel2017}): 

\newtheorem{theorem1}{Theorem}
\begin{theorem1}
 For every positive definite function $k(·, ·)$ on $\mathcal{X} \times \mathcal{X}$ there exists a unique RKHS with $k$ as its reproducing kernel. Conversely, the reproducing kernel of an RKHS is unique and positive definite.
\end{theorem1}

For RKHS the feature function can be derived as follow:

\vspace*{-4pt}
\begin{equation}
\phi(x) = k(\cdot, x)
\label{eq:03}
\end{equation}

This is referred to as a canonical feature map. 
Kernel methods that can be used with a feature mapping of distributions are needed for the ABC method.

\subsection{Kernel mean embedding}

Kernel methods have been extended from points transformation to distributions mapping \citealp{PatternRecognition2006}. 
Row data points $x \in R^d$ can be considered as distribution $P(x)$ in real space. 
The data points can be embedded in a feature space as a distribution in such a way that kernel mean embedding will represent the distribution in RKHS. 
For this purpose, the probability space is used as a triplet $\{ \Omega, \Sigma, P \}$, 
where $\Omega$ is the sample space, 
$\Sigma \subseteq 2^{\Omega}$ is sigma algebra with a subset of the power of $\Omega$, and
$P \in [0,1]$ is the probability measure on $\Sigma$ algebra 
\citealp{Wassermanbook2010}.

Kernel mean embedding based on a random variable can be described as a measurable function defined on a probability space $\{ \Omega, \Sigma, P \}$ that maps from the sample space $\Omega$ to the real numbers \citealp{Wassermanbook2010}. 
Let $X \subset R^d$ and $Y \subset R^d$ be two random variables such that $X: \Omega \to \mathcal{X}$ and $Y: \Omega \to \mathcal{Y}$. The probabilities defined on $X,Y$ may be marginal, joint, or conditional \citealp{KernelBayesRule2013}. 
Marginal probabilities $P(X)$ and $P(Y)$ are the (unconditional) probabilities of an event occurring;
joint probability $P(X,Y)$ is the probability of event $X = x$ and $Y = y$ occurring.
The conditional distribution $P(Y|X)$ shows the functional relationship between two random variables.

To produce a distribution mapping, \textbf{kernel mean embedding} is used with the $P$ probability measure in the form \citealp{ReviewKernel2017}: 

\vspace*{-10pt}

\begin{equation}
\begin{split}
\mu_P & := \mathbf{E}_P[\phi(X)]_\mathcal{H} \\ 
& = \int^{}_{\mathcal{X}} \phi(x) dP(x) \\
& = \int^{}_{\mathcal{X}} k(\cdot ,x) dP(x), 
\end{split}
\label{eq:04}
\end{equation}
where $k(\cdot ,x)$ is a kernel on $\mathcal{X}$, and $\mathbf{E}$ is expectation in feature space. 
If $P(x)$ is differentiable then $dP(x) = \rho(x)dx$ and $\rho(x)$ is the distribution density.
For a conditional distribution, this definition can be rewritten as:

\vspace*{-10pt}

\begin{equation}
\begin{split}
\mu_{P(\cdot|X)} & := \mathbf{E}_{P(\cdot|X)}[\psi(Y)|X]_\mathcal{G} \\ 
& = \int^{}_{\mathcal{Y}} \psi(y) dP(y|X) \\
& = \int^{}_{\mathcal{Y}} l(\cdot ,y) dP(y|X)
\end{split}
\label{eq:05}
\end{equation}
where $l(\cdot ,y)$ is a kernel on $\mathcal{Y}$.
To construct the Hilbert Space Embedding of a conditional distribution, the two positive definite kernels $k$ and $l$ for the $\mathcal{X}$ and $\mathcal{Y}$ domains, respectively, need to be defined: 

\vspace*{-10pt}

\begin{equation}
\begin{split}
& k: \mathcal{X \times X} \to R \qquad \qquad \text{ and } \qquad l: \mathcal{Y \times Y} \to R \\
& k(\cdot, x) = \phi(x) \in \mathcal{H}  \qquad \text{ and } \qquad l(\cdot, y) = \psi(y) \in \mathcal{G}
\end{split}
\label{eq:06}
\end{equation}
where $\mathcal{H}$ and $\mathcal{G}$ are the corresponding reproducing kernel Hilbert spaces. 
Based on the given sample $\{ x_1, x_2, \dots, x_n \}$ drawn from $P(x)$, the estimation of $\mu_P$ can be derived using the formula \citealp{ReviewKernel2017}

\vspace*{-6pt}
\begin{equation}
\mu_P  = \mathbf{E}_P[\phi(X)]_\mathcal{H}  
 \approx \frac{1}{n} \sum^{n}_{i=1}k(\cdot, x_i) \in\mathcal{H}
\label{eq:07}
\end{equation}

\vspace*{-4pt}
An important condition for kernels $k(\cdot ,x)$ and $l(\cdot ,y)$ is that they should be Bochner integrable or bounded \citealp{KernelBayesRule2013}.
In practice, one can use a \textbf{characteristic} kernel such that the kernel mean represents the information describing the entire distribution $P(x)$ \citealp{ReviewKernel2017}.

\subsection{Kernel Bayes rule}

The target of kernel ABC is to estimate the kernel mean of the conditional distribution. 
Having the kernel mean representation of a distribution allows us to establish Bayes' equality and to then transform it into the feature space \citealp{KernelBayesRule2013}:

\vspace*{-6pt}
\begin{equation}
P(Y|X) = \frac{P(X,Y)}{P(X)} = \frac{P(X|Y) \cdot P(Y)}{P(X)} 
\label{eq:08}
\end{equation}
Here for two random variables $X$ and $Y$ the conditional distributions are denoted as $P(Y|X)$ for all points of $X$ and $P(Y|X =x)$ for certain point $X = x$.
Using the same denotations as in Eq.~\ref{eq:06} 
the conditional mean embeddings $\mu_{P(Y|X)}$ and $\mu_{P(Y|X=x)} = \mu_{P(Y|x)}$ of these distributions are defined as expectations in RKHS
\citealp{ReviewKernel2017}:

\vspace*{-15pt}
\begin{equation}
\begin{split}
& \mu_{P(Y|X)}: \mathcal{H} \to \mathcal{G} \\  
& \mu_{P(Y|X)}  \in \mathcal{H} \times \mathcal{G}  \\ 
& \mu_{P(Y|x)} 
 = \mathbf{E}_{P(Y|x)}[\psi(Y)] \\
& \qquad \qquad \; \;  = \mu_{P(Y|X)} k(\cdot, x) 
\label{eq:09}
\end{split}
\end{equation}
Following these definitions $\mu_{P(Y|X)}$ is a matrix operator from $\mathcal{H}$ to $\mathcal{G}$ and $\mu_{P(Y|x)}$ is an element in $\mathcal{G}$. Thus, $\mu_{P(Y|x)}$ is the conditional expectation of the feature map of $Y$ given that $X = x$ can be represented as the marginal embedding. The embedding operator $\mu_{P(Y|X)}$ represents the conditioning operation that gives the result as $\mu_{P(Y|X=x)}$, i.e., one point from the entire set. 

To produce kernel mean embedding, the covariance operators are used in the form:

\vspace*{-16pt}
\begin{equation}
\begin{split}
& C_{YX} : \mathcal{H} \to  \mathcal{G} \\
& C_{YX} := \mathbf{E}_{P(Y,X)}[\psi(Y) \otimes \phi(X)] = \mu_{P(Y,X)} \\
& C_{XX} : \mathcal{H} \to  \mathcal{H} \\
& C_{XX} := \mathbf{E}_{P(X)}[\phi(X) \otimes \phi(X)] 
\label{eq:10}
\end{split}
\end{equation}
where $C_{YX}$ is a cross-covariance operator, 
$C_{XX}$ is a covariance operator, 
and $\otimes$ is the outer product of two vectors, 
$\psi(Y) \otimes \phi(X) \equiv  \psi(Y) \times \phi^\mathsf{T}(X)$, 
that gives the result as a matrix with elements 
$[\psi(Y) \times \phi^\mathsf{T}(X) ]_{ij} = \psi(y_i) \times \phi(x_j)$.
From these definitions the Kernel mean embedding can be represented as follows (Definition 4.1 in \citealp{ReviewKernel2017}):

\vspace*{-16pt}
\begin{equation}
\begin{split}
& \mu_{P(Y|X)} = C_{YX} C^{-1}_{XX} \\
& \mu_{P(Y|X=x)} = C_{YX} C^{-1}_{XX} k(\cdot, x)
\label{eq:11}
\end{split}
\end{equation}

The most important theorem for our research is the following (Theorem 4.1 in \citealp{ReviewKernel2017}):

\begin{theorem1}
Let $\mu_{\pi} \in \mathcal{H}$ and $\mu_{Q_y} \in \mathcal{G}$ be the kernel mean embeddings of $\pi$  and $Q_y$ distributions, respectively.
Let $R(C_{XX} )$ denote the range space (span of the column vectors) of a covariance operator $C_{XX}$.
If $C_{XX}$ is injective, $\mu_{\pi} \in R(C_{XX})$, and $\mathbf{E}[\psi(Y)] \in \mathcal{G}$ for any $\psi \in \mathcal{G}$, then
\end{theorem1}

\vspace*{-6pt}
\begin{equation}
\begin{split}
& \mu_{Q_y} = C_{YX} C^{-1}_{XX} \mu_{\pi}, \\
& \mu_{\pi} \in \mathcal{H} \\ 
& \mu_{Q_y}  \in \mathcal{G} 
\label{eq:12}
\end{split}
\end{equation}

In practice the joint distribution $P(X,Y)$ is unknown, which is the reason that $C_{XX}$ and $C_{YX}$ cannot be computed directly. 
Empirical estimation of $\mu_{Q_y}$ is based on the sample $(x_1, y_1), ... , (x_n, y_n)$ from $P(X,Y)$, which arises from the independent and identically distributed random variables $X$ and $Y$. 
Then, the conditional mean embedding $\mu_{P(Y|X=x)}$ can be estimated using the following (Theorem 4.2 in \citealp{ReviewKernel2017}):

\vspace*{-6pt}
\begin{equation}
\begin{split}
& \mu_{P(Y|X=x)} \approx \Psi (\mathbf{G} + n \lambda \mathbb{I}_n)^{-1} \Phi^\mathsf{T} k(\cdot, x) \\
& \Psi^\mathsf{T} := [\psi(y_1), \psi(y_2), ..., \psi(y_n)]^\mathsf{T}  \\
& \Phi^\mathsf{T} := [\phi(x_1), \phi(x_2),...,\phi(x_n)]^\mathsf{T} \\
& \mathbf{G} = \Phi^\mathsf{T} \Phi
\label{eq:13}
\end{split}
\end{equation}
where $\mathbf{G}$ is a Gram matrix in $\mathcal{H}$, $\mathbb{I}_n$ is the identity matrix of rank $n$, and $\lambda$ is a positive regularization constant. 

\subsection{Kernel ABC}

In rejection ABC, a large number of simulations  are iteratively generated to produce the posterior distribution. The summary statistics for each simulation reduce the dimension of the problem. Finally, a decision is made regarding the rejection or acceptance of a simulation \citealp{KernelRecursiveABC2018}; \citealp{ABCPrictice2010}. The accepted simulations provide the posterior distribution of parameters, and the maximum a posterior is typically used as parameter estimation for the given observation.
In kernel ABC, an element of RKHS space represents a distribution, so, in principle, there is no need to use summary statistics. 
The maximum mean discrepancy $\mathbf{MMD}$ criterion is used to check the consistency of the distributions of the observation and simulation data instead \citealp{K2-ABC2016}.
The target of kernel ABC is to achieve kernel mean embedding of the posterior distribution using Eq.~(\ref{eq:13}), and then generate a sampling using the complementary algorithm such as kernel herding \citealp{KernelHerding2010}. Finally, parameter estimation can be accomplished based on the posterior distribution of the new sampling.
To carry out kernel ABC, the definition of the model, observation and simulation data, and RKHS are needed.

Let us assume a stochastic model that has measurable variable $\Theta \in \mathcal{Q}$ as an input $d$-dimensional parameter and that the results of the model's simulation are represented by another measurable variable $S \in \mathcal{S}$ of $d_s$ dimensions.
$\mathcal{Q}$ and $\mathcal{S}$ are the corresponding domains. 
Note that the result is not deterministic due to the stochastic nature of the simulation. 
Let us further consider observation point $s^* \in \mathcal{S}$ and
a given sample of $n$ simulations $S = \{ s_{1}, s_{2}, \dots  , s_{n}\} \in \mathcal{S}$ based on related input parameters  $\Theta = \{ \theta_1,...,\theta_n\} \in \mathcal{Q}$.
The corresponding prior distributions of parameters $\pi(\theta)$ and simulations $P(s)$ are given by:

\vspace*{-6pt}
\begin{equation}
\begin{split}
& S = \{ s_1, s_2, ..., s_n  \} \in \mathcal{S} \\
& \Theta = \{ \theta_1,...,\theta_n\} \in \mathcal{Q} \\
& s \sim P(s) \\
& \theta  \sim  \pi(\theta)
\label{eq:14}
\end{split}
\end{equation}

Therefore, the posterior distribution $P(\theta|s^*)$ of the parameters $\theta$ on given observation $s^*$ can be obtained from Eq.~(\ref{eq:08}) in the form:

\vspace*{-6pt}
\begin{equation}
P(\theta|S=s^*) \propto P(s^* | \theta) \cdot \pi(\theta),
\label{eq:15}
\end{equation}
where $P(s,\theta)$ is the joint distribution of $s$ and $\theta$ and $P(s | \theta)$ is the conditional distribution. 
In kernel ABC, the kernel mean embedding of the posterior distribution $P(\theta|S=s^*)$ is defined as follows \citealp{K2-ABC2016} and \citealp{KernelRecursiveABC2018}:

\vspace*{-6pt}
\begin{equation}
\mu_{P(\theta|S = s^*) } = \int_{\mathcal{Q}} l(\cdot,\theta) dP(\theta|s^*)   \in \mathcal{G}
\label{eq:16}
\end{equation}
where $l(\cdot,\theta)\in \mathcal{G}$ denotes the kernel in reproducing kernel Hilbert space $\mathcal{G}$. In order to exploit Eq.~(\ref{eq:16})  one must define two kernels: $l(\cdot, \theta)$ for the domain of parameters $\mathcal{Q}$ and $k(\cdot, s)$ for the domain of simulations $\mathcal{S}$:

\vspace*{-6pt}
\begin{equation}
\begin{split}
& k: \mathcal{S \times S} \to R \qquad  \qquad \text{ and } \qquad l: \mathcal{Q \times Q} \to R \\
& k(\cdot, s) = \phi(s) \in \mathcal{H}  \qquad \text{ and } \qquad l(\cdot, \theta) = \psi(\theta) \in \mathcal{G},
\label{eq:17}
\end{split}
\end{equation}
where $\mathcal{H}$ is the reproducing kernel Hilbert space for $k(\cdot, s)$, and $\phi(s)$ and $\psi(\theta)$ are the mapping functions for $\mathcal{H}$ and $\mathcal{G}$ respectively.
Using the same notation, we can rewrite Eq.~(\ref{eq:13}) replacing $X,Y$ with $S,\Theta$ and $x, y_i$ with $s^*,\theta_i$ \citealp{ReviewKernel2017}:

\vspace*{-10pt}
\begin{equation}
\begin{split}
\mu_{P(\theta|S=s^*)} 
& \approx \Psi (\mathbf{G} + n \lambda \mathbb{I}_n)^{-1} \Phi^\mathsf{T} k(\cdot, s^*), \\
& \Psi^\mathsf{T} := [\psi(\theta_1), \psi(\theta_2), ..., \psi(\theta_n)]^\mathsf{T}  \\
& \Phi^\mathsf{T} := [\phi(s_1), \phi(s_2),...,\phi(s_n)]^\mathsf{T} \\
& \mathbf{G} = \Phi^\mathsf{T} \Phi
\label{eq:18}
\end{split}
\end{equation}

To illustrate the numerical calculation with $\Psi$ and $\Phi$ 
let us define the dimensions of the Hilbert spaces $\mathcal{H}$ and $\mathcal{G}$ as $d_\phi$ and $d_\psi$.
Then, Gram matrix $G \in R^{n \times n}$ can be represented as an $n \times n$ matrix, 
$\Psi \in R^{d_\psi \times n}$ can be represented as a $d_\psi \times n$ matrix
and $\Phi$ is similar.
Thus, the Gram matrix can be explicitly rewritten as follows \citealp{ReviewKernel2017} and \citealp{Featurespace2021}: 

\begin{equation}
\mathbf{G}:= 
\begin{pmatrix}
k(s_1, s_1) & \cdots & k(s_1, s_n) \\
\vdots &  \ddots  & \vdots \\
k(s_n, s_1) & \cdots & k(s_n, y_n) 
\end{pmatrix} \in R^{n \times n}
\label{eq:19}
\end{equation}

Because $k(\cdot, s^*) = \phi( s^*) \in R^{d_\phi}$ is a $d_\phi \times 1$ matrix and
$\Phi^\mathsf{T} \in R^{n \times d_\phi}$ can be represented as an $n \times d_\phi$ matrix,
the result of $\Phi^\mathsf{T} k(\cdot, s^*)$ is an $n \times 1$ matrix \citealp{Featurespace2021}:

\vspace*{-6pt}
\begin{equation}
\begin{split}
\Phi^\mathsf{T} k(\cdot, s^*)
& = [\phi(s_1), \phi(s_2), \dots ,\phi(s_n)] \phi( s^*) \\
& = [\phi(s_1)\phi( s^*), \phi(s_2)\phi( s^*), \dots ,\phi(s_n)\phi( s^*)]^\mathsf{T}  \\
& = [ k(s_1,s^*), k(s_2, s^*), \dots , k(s_n, s^*) ]^\mathsf{T} \in R^{n \times 1} 
\label{eq:20}
\end{split}
\end{equation}

Moreover, $(\mathbf{G} + n \lambda \mathbb{I}_n)^{-1} \in R^{n \times n}$ is an $n \times n$ matrix; 
therefore, the result of $\mathbf{w}  =  (\mathbf{G} + n \lambda \mathbb{I}_n)^{-1} \Phi^\mathsf{T} k(\cdot, s^*) = (w_1, w_2, \dots, w_n)^\mathsf{T} \in R^{n \times 1}$ is represented as the weights in an $n \times 1$ matrix.
Usually, Eq.~(\ref{eq:18}) is written as the sum of products of the weights $\mathbf{w}$ and feature mapping $l(\cdot,\theta_i) = \psi(\theta_i), i \in [1,n]$ \citealp{ReviewKernel2017}:

\vspace*{-10pt}
\begin{equation}
\begin{split}
\mu_{P(\theta|S=s*)} 
& \approx \Psi (\mathbf{G} + n \lambda \mathbb{I}_n)^{-1} \Phi^\mathsf{T} k(\cdot, s^*) \\
& = \Psi \times \mathbf{w} \\
& = \sum_{i=1}^n w_i \psi(\theta_i)  \in \mathcal{G}
\label{eq:21}
\end{split}
\end{equation}

Under the conditions $n \to \infty$ and $\lambda \to 0$, the estimation will consistently converge to the true value \citealp{KernelBayesRule2013} and \citealp{ReviewKernel2017}.

\vspace*{32pt}


\vspace*{-12pt}
\section{ABC based on Maxima Weighted Isolation Kernel Mapping}

\subsection{Isolation Kernel}

Isolation Kernel is based on the Isolation Forest (iForest) algorithm \cite{IsolationForest2008} (see details in Application 1).
The formal definition of Isolation Kernel is as follows \cite{iKernel_SVM2018}:

\begin{itemize}

\item  Let $D = \{ x_1,...,x_n \}$ , $x_k \in R^d$ be a dataset sampled from an unknown probability density function $x_k \sim \mathcal{P_D}$. 

\item  Let $H_\xi (D)$ denote the set of all partitions $H$ that are admissible from the dataset $\mathcal{D} \subset D$ where each point $z \in \mathcal{D}$ has an equal probability of being selected from $D$; and $|\mathcal{D}| = \xi$. In other words, subset $\mathcal{D} = \{z_1, z_2, ..., z_{\xi} \}$ was extracted from initial set $D = \{ x_1,...,x_n \}$ ($\xi \ll n$) in order to construct partitions $H_\xi (D)$.

\item  Each isolating partition $\eta[z] \in H_{\xi}$ isolates one point $z \in \mathcal{D}$ from the rest of the points in a random subset $\mathcal{D}$, where
$\xi$ is the number of elements in the subset $\mathcal{D}$. 

\item Definition 2.1 in \cite{iKernel_dist2020} established the following: 

\newtheorem{definition1}{Definition}
\begin{definition1}
For any two points $x, y \in R^d$ , Isolation Kernel of ${x}$ and ${y}$ is defined to be the expectation taken over the probability distribution on all partitionings $H \in H_\xi (D)$ that both ${x}$ and ${y}$ fall into the same isolating partition $\eta[z] \in H$, where $z \in \mathcal{D}, \xi = |\mathcal{D}|$:
\end{definition1}
\end{itemize}

\vspace*{0pt}
\begin{equation}
\begin{split}
k_I (x,y | D) & = \mathbb{E}_{H_{\xi}(D)} [ \mathbf{1}(x,y \in \eta[z] | \eta[z] \in H)  ]  \\
&  = \mathbb{E}_{\mathcal{D} \in D}[ \mathbf{1}(x,y \in \eta[z] | z \in \mathcal{D})  ]  \\
&  = P( x,y \in \eta[z] | z \in \mathcal{D} \subset D ), 
\label{eq:22}
\end{split}
\end{equation}
where $\mathbf{1}(B)$ is the indicator function that outputs $1$ if $B$ is true; otherwise, $\mathbf{1}(B) = 0$.
Equation (4) in \cite{iKernel_dist2020} shows how to calculate Isolation Kernel in practice using an approximation for a finite number of partitionings:

\vspace*{0pt}
\begin{equation}
\begin{split}
 k_I (x,y | D) 
 & \approx \frac{1}{t} \sum^{t}_{i=1} \mathbf{1}(x,y \in \eta | \eta \in H_i)  \\ 
&  \approx \frac{1}{t} \sum^{t}_{i=1} \sum_{\eta \in H_i} \mathbf{1}(x \in \eta)\mathbf{1}(y \in \eta)  \\ 
&  \approx \frac{1}{t} \Phi^\mathsf{T}(x) \times \Phi(y),
\label{eq:23}
\end{split}
\end{equation}
where $\Phi(x)$ is a feature mapping of $x$. 
In accordance with Definition 3.1 in \cite{iKernel_dist2020}, the feature mapping $\Phi(x): x \to \{ 0,1 \}^{t \times \xi}$ of $k_I$ is a vector that represents the partitions in all the partitioning $H_i \in \mathbb{H}_{\xi}(D)$, $i = [1,t]$, where $x$ falls into only one of the $\xi$ partitions in each partitioning $H_i$.

The accuracy of the approximation is estimated for the iTree method of partitioning in \cite{iKernel_SVM2018} and is represented as approximating the probability distribution of two points of $\delta$-distance falling into the same isolating partition: 

\vspace*{0pt}
\begin{equation}
k_I(x, \delta) \approx \xi^{-\varkappa \delta },
\label{eq:24}
\end{equation}
where $\varkappa \le 0$ is an integer indicating the multiplies of $\delta$ from $x$. The $\xi$ plays the role of sharpness; by an increase of $\xi$, the resolution of Isolation Kernel is also improved. The upper boundary for $\xi$ is the condition $\xi \ll n$ in order to maintain a diversity of partitionings. The final parameter $t$ can be chosen based on computational cost, recognizing that, while increasing the number of trees can improve accuracy, it also increases the computation time.  

Isolation Kernel is positive definite in quadratic form and defines an RKHS $\mathcal{H}$ \cite{iKernel_dist2020}. 
Various isolation partitioning mechanisms can be used to implement Isolation Kernel, including iForest, the Voronoi diagram \cite{iKernel_SVM2018}, and a newly proposed method based on hyperspheres \cite{iKernel_dist2020}. 
iForest requires choosing several issues in an iteration: 
First, a subset $\mathcal{D}$ of points $\{ z_1, ..., z_{\xi} \}$ must be determined; 
then, a dimension with a point for the splitting of multidimensional space $R^d$ must be chosen; 
finally, a large number of $iTrees$ must be constructed.
Then, the time consumed in the training stage is proportional to 
$O(\xi \times \log_2\xi \times t)$. 
The evaluation stage is needed to compute the length of the $iTree$ for each data point $x \in D$ using recursive function \cite{iKernel_dist2020}. 
In this case the computational cost will be $O(\log_2\xi \times t)$ for each new point. 

\begin{figure}[!tpb]
\centerline{\includegraphics[width=85mm]{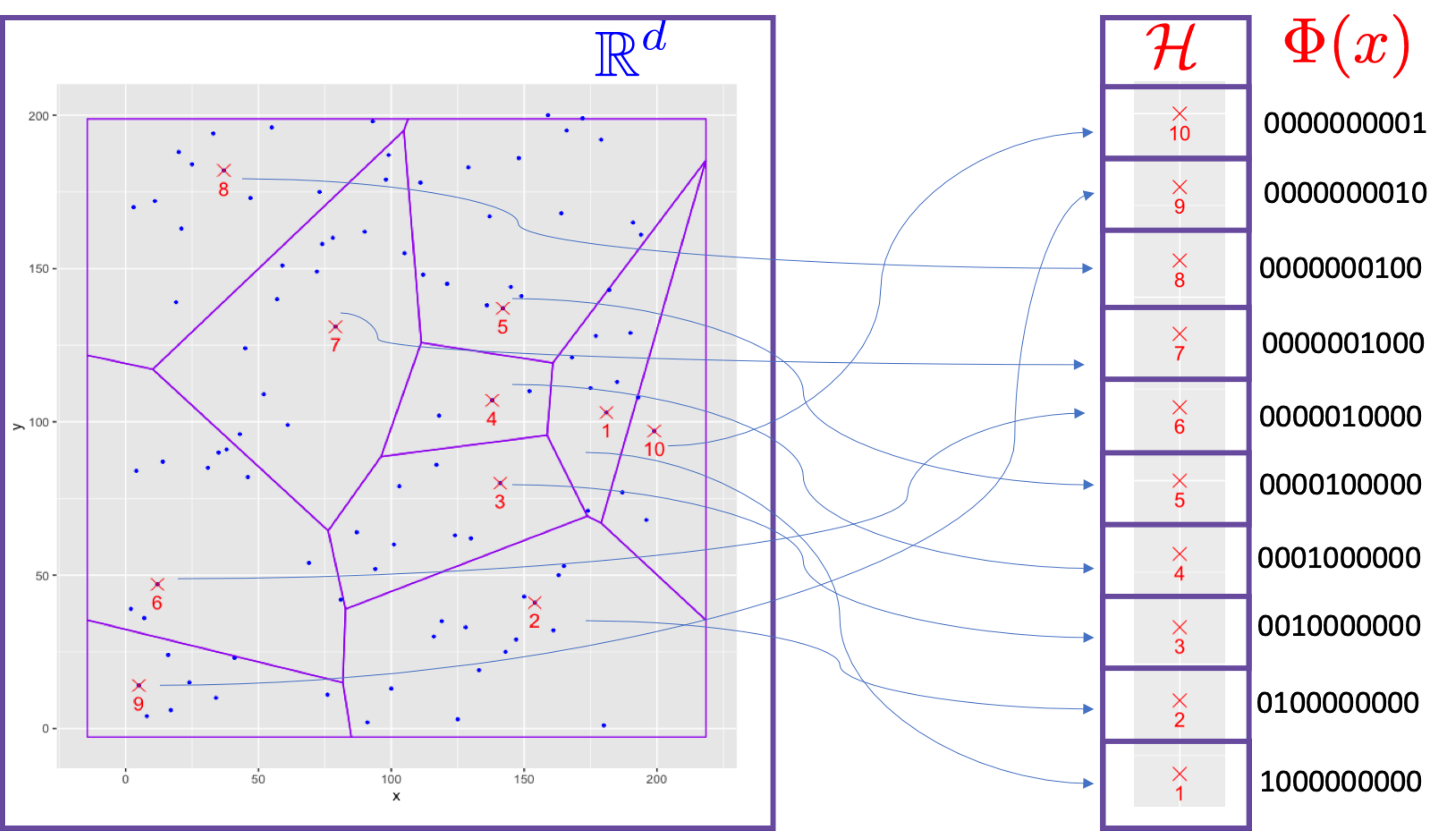}}
\caption{An illustration of feature map $\Phi(x)$ of Isolation Kernel for the partitioning of ten partitions. Each partition relates to a point (shown as a red cross) $z \in \mathcal{D}$ where $|\mathcal{D}| = \xi = 10$ are randomly selected from the given dataset $D$. When a point $x_i$ is in partition region $\eta[z_j]$, $x_i$ is regarded as being at the related point in $\mathcal{H}$ space, and the coding by feature map $\Phi(x_i)$. $\Phi(x_i)$ is a binary vector with a single `1' whose position corresponds to the partition number.}
\label{fig:02}
\end{figure}

\subsection{Isolation Kernel based on Voronoi diagram}

Let's now consider the algorithm of Isolation Kernel based on the Voronoi diagram \cite{iKernel_dist2020}.
In the simplest two-dimensional case, a Voronoi diagram is a partition of a plane into regions close to each of a given set of points (called Voronoi sites or seeds) \cite{Voronoi1987}, \cite{Voronoi2019}. For each site, there is a corresponding region, called a Voronoi cell, consisting of all points of the plane closer to that site than to any other \cite{Voronoi1984}, \cite{Voronoi2020}. Fig.~\ref{fig:02} shows an example of partitioning using a Voronoi diagram. Voronoi sites are indicated by the numbered red crosses. The Voronoi cells are represented as polygons dividing the entire plane. The advantage of the Voronoi diagram is that it offers very simple way of dividing all points into cells merely by calculating distances and choosing the smallest of these distances.

The algorithmic steps of feature mapping $\Phi(x)$ using a Voronoi diagram are presented in Algorithm~\ref{alg:Voronoi}; the data transformation is shown in Fig.~\ref{fig:02}.
The algorithm calculates feature mapping $\Phi(x)$ for each point $x$ in dataset $D = \{x_1, x_2, ..., x_n \} \in R^d$, where $d$ is the dimension and $t$ is the number of partitionings. 

\begin{algorithm}[tbh]
 \KwIn{Dataset $D = \{x_1, x_2, ..., x_n \} \in R^d$ } 
 
  \For{i = 1 .. t}{
\Begin { {\bf Create partitioning} $\mathcal{H}_i$:\\
Randomly produce subset $\mathcal{D}_i \subset D$, where $\mathcal{D}_i = \{z_{i1}, z_{i2}, ..., z_{i \xi} \}$ \\
Construct a Voronoi diagram with partitions $\mathcal{H}_i$ and \\ 
corresponding Voronoi cells $\{V_{ik}(z_{ik}) \}, k \in [1,\xi]$\\
}
\ForEach{$x \in D$ }
{  \ForEach{$\eta_j \in \mathcal{H}_i$, $j = 1 .. \xi$ }{
Calculate $\Phi_{ij}(x) = {\bf 1}(x \in \eta_j | \eta_j \in \mathcal{H}_i)$ \\
}
Calculate $\Phi_i(x) = [0000...1...000]$, where the position of $1$ corresponds to Voronoi cell $\upsilon_i(x) \in \{V_{ik}(z_{ik}) \}$ to which $x$ belongs \\
}
} 
\KwResult{$\Phi (x) =  \textrm{concatenation } \{ \Phi_1(x), ..., \Phi_t(x) \} $}
 \KwOut{$\{ \mathcal{D}_i \}, \{ \mathcal{H}_i \}, \{ \upsilon_i(x) \}, \{ V_{ik} ( z_{ik} ) \}, i \in [1,t], k \in [1,\xi]$}
 \label{alg:Voronoi}
 \caption{Algorithm to calculate feature mapping $\Phi (x)$}
\end{algorithm}
 
For each partitioning $\mathcal{H}_i$ the number of partitions is fixed at $\xi$ and the value of feature mapping for each point $x$ is $\Phi_i(x)$ which is a binary vector with only one 1 and zeros in the other positions to represent the fact that $x$ in only one partition in each partitioning $H_i$. 
The feature mapping is achieved by concatenation of all $\Phi_i(x)$.

\vspace*{0pt}

The partitioning is implemented as follows: 
First, choose from original dataset $D$ the sites $z_j \in \mathcal{D}, j=[1..\xi]$ that define partitions $\eta[z_j]$ (Fig.~\ref{fig:02}). 
For each point $x_i \in D$ we can determine to which partition the point $x_i$ belongs by calculating the distances to all points $z \in \mathcal{D}$ and choosing the smallest of the distances. In this manner we are able to separate all points $x_i$ and reflect them into feature map $\Phi(x_i)$ as a binary number with a single `1' at the position corresponding to the partition number. 
This mechanism has been shown to produce large partitions in sparse regions and small partitions in dense regions \cite{iKernel_dist2020}.

The Voronoi diagram requires a subset of $\xi$ points for $t$ trees at the training stage, so that the computational cost is evaluated as $O(\xi \times t)$.
At the evaluation stage, for each new point $x$, it is necessary to find the nearest point from $x$ to one point of subset $\mathcal{D} = \{ z_1, ..., z_{\xi} \}$, and thus it is necessary to find $\xi$ distances. Making this calculation for each tree, we can estimate the time consumed as $O(\xi \times t)$, which is of the same order of cost as in the training stage.

\subsection{Maxima weighted iKernel mapping}

To produce the posterior distribution for parameters $P(\theta|S = s^*)$ under given observation $s^*$, it is necessary to generate sampling of $\theta$ from 
$\mathcal{Q}$ Eqs.~(\ref{eq:15} and \ref{eq:16}). 
A number of appropriate methods to generate such sampling are available, including Kernel Herding \citealp{ParametricHerding2010}, Markov Chain Monte Carlo (MCMC) \citealp{ARandomPolynomial1991}, Sequential Monte Carlo (SMC) \citealp{SMC1998}, Sequential Kernel Herding \citealp{SequentialKernelHerding2015}.

Note that the dimensions $d_\phi$ and  $d_\psi$ of Hilbert spaces $\mathcal{H}$ and $\mathcal{G}$ can be infinite in principle, which is the reason that the kernel trick is used in computations to avoid the direct calculation of kernel mean embedding. 
A common approach to checking convergence is to use the maximum mean discrepancy  $\mathbf{MMD}$, which also employs the kernel trick \citealp{MMD2006} and \citealp{K2-ABC2016}. 
To produce the initial and generated sampling, weights such as $\Phi^\mathsf{T} k(\cdot, s^*)$ are used. These, too, are based on the kernel trick Eqs.~(\ref{eq:20} and \ref{eq:21}).


As an alternative to the kernel trick method, we suggest the calculation of the maxima weighted state of the kernel mean embedding 
$\mu_{P(\theta|S=s*)}$ in RKHS $\mathcal{G}$ explicitly using Eq.~\ref{eq:21} with a transformation based on properties of Isolation Kernel (Eqs.~\ref{eq:22} and \ref{eq:23}) and the output of Algorithm~\ref{alg:Voronoi}.
The most important and interesting idea is that we produce  $\mu_{P(\theta|S=s*)}$ in RKHS $\mathcal{G}$ related to the space of parameters, which means that iKernel will give us information about the parameters related to the observation directly. 

Let us consider the properties of iKernel from its definition:

\begin{itemize}
\item iKernel mapping consists of numerous elements, each of which originates from the Voronoi diagram.

\item  Fig.~\ref{fig:02} shows that each element in RKHS $\mathcal{G}$ identifies a Voronoi cell from the entire diagram.

\item The number of elements in RKHS $\mathcal{G}$ is determined by the approximation in Eq.~\ref{eq:23} and relates to the number of trees $t$ in the Isolation Forest or the number of Voronoi diagrams. 

\end{itemize}

Maintaining the notation of Eq.~(\ref{eq:14}) and in order 
to prepare iKernel, we can define Voronoi diagrams using a sample of parameters $\Theta$ with size $n$ for each tree in accordance with Algorithm~\ref{alg:Voronoi}: 

\vspace*{0pt}
\begin{equation}
\begin{split}
& z_{jk} \subset \Theta \qquad j \in [1,t] \qquad k \in [1, \xi] \text{  and  } \xi \ll n  \\  
& V_{jk} (z_{jk}) = \{   \theta \in \mathcal{Q} | d(\theta, z_{jk} ) \le  d(\theta, z_{pk} ) , p \ne j \},
\label{eq:25}
\end{split}
\end{equation}
where $ z_{jk}$ is the $k$-th site of the $j$-th tree or Voronoi diagram and element of a sample of parameters $\Theta$, $\xi$ is a number of sites in each Voronoi diagram/tree, $t$ is the number of trees/Voronoi diagrams, $V_{jk} (z_{jk})$ is the Voronoi cell related to site $z_{jk}$, and $d( \cdot, \cdot )$ is a distance metric.

Next, we can define Isolation Kernel and feature mapping $l(\cdot,\theta) = \psi(\theta)$ related to Algorithm~\ref{alg:Voronoi} as follows:

\vspace*{0pt}
\begin{equation}
\begin{split}
& V_{jk} (z_{jk})  \xrightarrow[]{Alg.\ref{alg:Voronoi}}   \psi (\theta)  \\ 
& \psi (\theta_i)  \overset{ Alg.\ref{alg:Voronoi}}{\Longleftarrow \Longrightarrow} \big \{ \zeta_{i1}, ..., \zeta_{it} \big \} =  \big \{ \zeta_{ij} \big \}  \\   
& \zeta_{ij}  =  \underset{k \in [1,\xi] }{\arg\min} \quad d( \theta_i, z_{jk} ),
\label{eq:26}
\end{split}
\end{equation}
where $\zeta_{ij}$ is the Voronoi site with minimal distance to the $\theta_i$ parameter from the set of sites $\big \{  z_{jk}  \big \}$ of the $j^{th}$ Voronoi diagram/tree.

Eq.~(\ref{eq:21}) gives information regarding the contribution of each sample point to the kernel mean embedding of the observation point in RKHS $\mathcal{G}$.  The weights $\Phi^\mathsf{T} k(\cdot, s^*)$ are similarity coefficients between sample points and the observation point. 
Thus, the series of the sum in Eq.~(\ref{eq:21}) gives information about the Voronoi diagrams in the parameters space through the similarity between the sample points and the observation point in the space of $S$ from results of simulations. 
Incorporating Eqs.~(\ref{eq:21},\ref{eq:25} and \ref{eq:26}) together enables us to describe iKernel mean embedding $\mu_{P(\theta|S=s*)}$ as a weighted distribution of Voronoi sites for each $j^{th}$ tree:

\begin{equation}
\begin{split}
& P_{jk}(z_{jk})  \Longleftrightarrow   \mu_{P(\theta|S=s*)}  = \sum_{i=1}^n w_i \psi(\theta_i)\\
& P_{jk}(z_{jk}) \propto \frac{1}{n} \sum_{i=1}^n w_i \times \mathbf{1}( \zeta_{ij} = z_{jk}), 
\label{eq:27}
\end{split}
\end{equation}
where $P_{jk}(z_{jk})$ is the probability of finding a parameter related to observation $s^*$ close to the $k^{th}$ Voronoi site $z_{jk}$ of the $j^{th}$ tree/diagram.

We now turn to an algorithm for seeking the parameter related to observation $s^*$ using the maxima weighted state of 
$\mu_{P(\theta | S=s*)}$. 
This procedure reduces the information from iKernel mean $\mu_{P(\theta | S=s*)}$ but allows for the possibility of finding a position in multi-dimensional space of parameters with high probability related to $s^*$. 
In this way, our strategy is based on the obvious hypothesis that the true value is near a point related to the maxima weighted iKernel mapping. 
For this purpose let us determine the maxima weighted probabilities $P_{jk}(z_{jk})$ across the trees:

\begin{equation}
\begin{split}
& z_{j}^*  =   \underset{k \in [1,\xi] }{\arg\max} \quad P_{jk}( z_{jk} )  \\
&  \upsilon_j^*  =  \upsilon_j ( z_{j}^* )  \\  
& \mu_{\upsilon^*}    \overset{ Alg.\ref{alg:Voronoi}}{\Longleftarrow \Longrightarrow}  \big \{  z_{1}^*, ... , z_{t}^* \big \},
\label{eq:28}
\end{split}
\end{equation}
where $\{ \upsilon^*_1, ... \upsilon^*_t \}$ and $\{ z^*_1, ..., z^*_t\}$ are Voronoi cells and sites related to the maxima weighted state of $\mu_{P(\theta | S=s*)}$ (denoted as $\mu_{\upsilon^*}$) of the observation point $s^*$ in the parameter space $\mathcal{Q}$.

After producing the maxima weighted iKernel mapping $\mu_{\upsilon^*}$ in the explicit form of the list of Voronoi cells of trees as well as an element of Hilbert space $\mathcal{G}$, it is possible to detect with high probability the position or tiny area $\upsilon^{*}$ in the parameter space related to the observation as an intersection of all the Voronoi cells:

\vspace*{0pt}
\begin{equation}
\begin{split}
& \upsilon^{*}  =  \bigcap_{j=1}^{t} \upsilon^*_j
\end{split}
\label{eq:29}
\end{equation}
We should note that $\upsilon^{*} $ can, in principle, be the empty set which is why we do not use the intersection area for calculation; 
we use it here simply to demonstrate the algorithmic features of our method and to show an example of a two-dimensional case.

\begin{figure}[t!]
\centerline{\includegraphics[width=80mm]{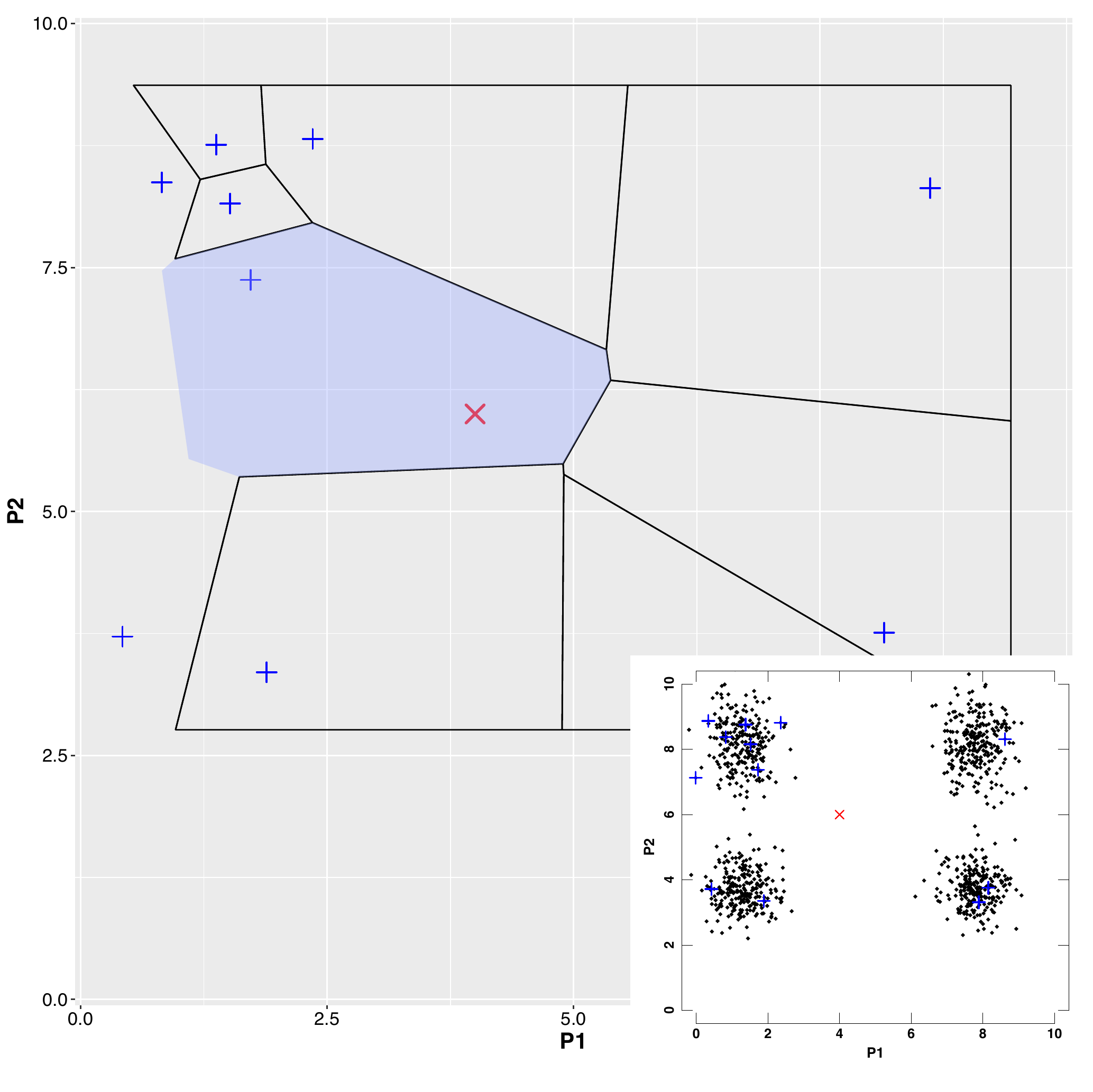}}
\caption{Voronoi diagram in the parameter space of $P1$ and $P2$: straight crosses ($+$) are Voronoi sites, the partitions are related to Voronoi cells, the oblique cross ($\times$) is the observation point and the shaded area is the Voronoi cell to which the observation belongs. \\
{\bf Insert:} Points in the sample dataset; the straight crosses (+) are a subset of Voronoi sites and the oblique cross ($\times$) is the observation point.}
\label{fig:03}
\end{figure}

\begin{figure}[!tpb]
\centerline{\includegraphics[width=87mm]{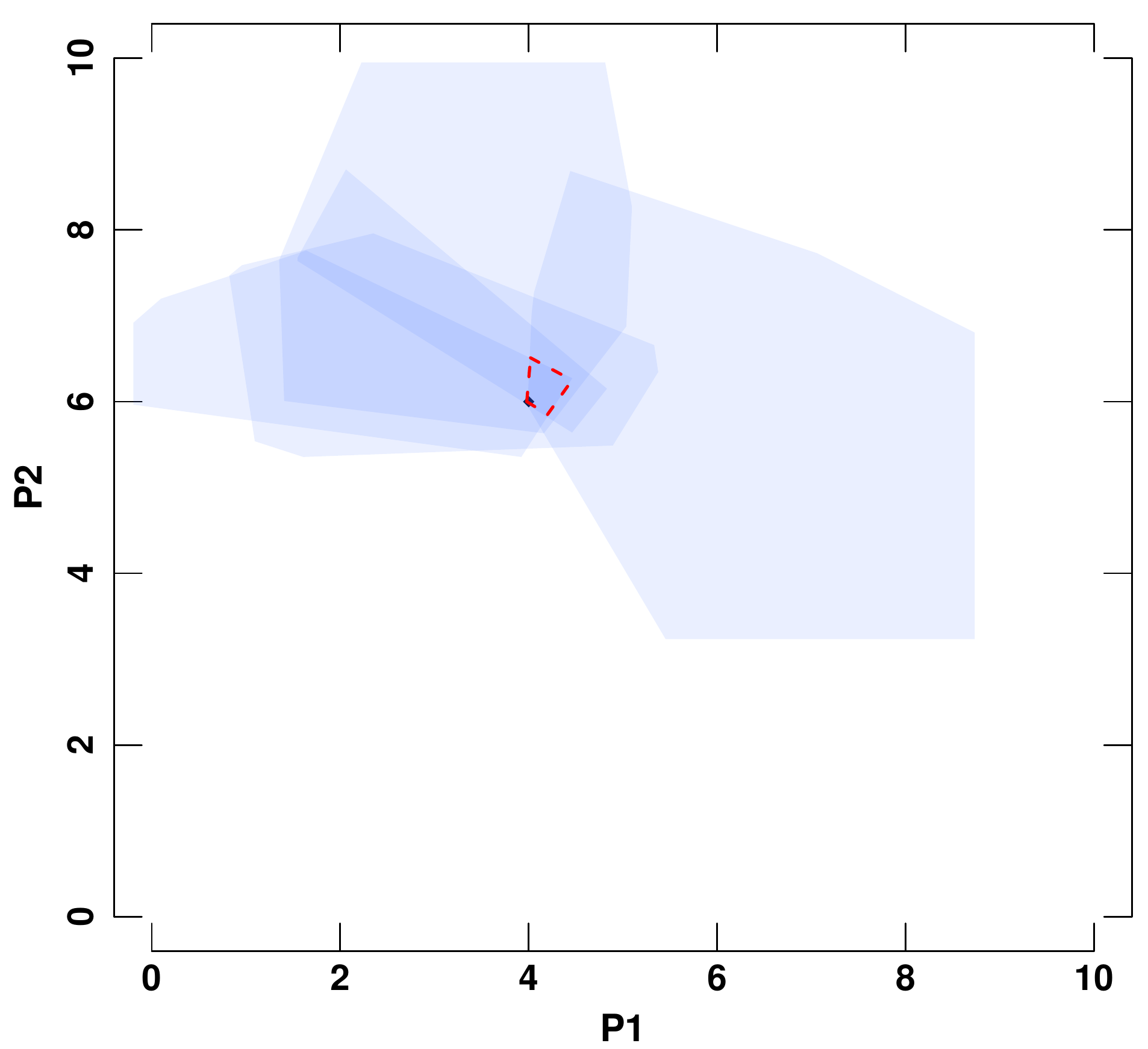}}
\caption{Intersection of the Voronoi cells from different Voronoi diagrams. Voronoi cells are shown as semi-transparent areas;
the intersection is enclosed by a dashed line.}
\label{fig:04}
\end{figure}

To explain the use of Eqs.~(\ref{eq:27}, \ref{eq:28} and \ref{eq:29}) of Isolation Kernel, let us consider an example with unevenly distributed sample data (insert in Fig.~\ref{fig:03}) of two-dimensional parameters ($P1$ and $P2$) and only twelve sites $\xi = 12$ with only five trees $t=5$.
A similar unevenness of the sample data appears under stochastic simulation with frequent null-output in a certain area of the parameters space.
One may call these `rare events', which commonly attract interest.
We insert the truth observation point into the plots to show the closeness that enables the algorithm to make a decision (oblique cross in Fig.~\ref{fig:03} and point in Fig.~\ref{fig:04}). 

Figure~\ref{fig:03} shows us an example of the Voronoi diagram and the Voronoi cell to which the observation point belongs.
Iteratively, the applied algorithm finds all Voronoi cells, and, ultimately, it is possible to find the intersection $\upsilon^*$ (Eq.~\ref{eq:29}) of all Voronoi cells from all Voronoi diagrams (Fig.~\ref{fig:04}).
As shown, the intersection $\upsilon^*$ of the Voronoi cells is a very small area, even with five cells or trees. In many calculations, we usually used 10-70 sites and 20-400 trees which are much more than in the example. 
In the supplementary video, the iterative process is shown for the selection of sites, construction of the Voronoi diagram, and determination of the intersection. 

The machine learning algorithm of ABC based on maxima weighted iKernel mapping $\mu_{\upsilon^*}$ consists of the following steps:

\begin{itemize}

\item Define the kernel and feature mapping $k(\cdot, s) = \phi(s) \in \mathcal{H} $ in the space of the results of simulation $\mathcal{S}$ Eqs.~(\ref{eq:14}, \ref{eq:17}). Note that any type of kernel with the reproducing property is acceptable (i.e., it does not need to be iKernel).

\item  Use Eqs.~(\ref{eq:19},\ref{eq:20}) to determine the Gram matrix $\mathbf{G}$ and weights $\Phi^\mathsf{T} k(\cdot, s^*)$ of similarity between the simulation and the observation points. 

\item {\bf Training step:} Execute Algorithm~\ref{alg:Voronoi} with iKernel $l(\cdot, \theta) = \psi(\theta) \in \mathcal{G}$ (Eq.~\ref{eq:17}) to produce Voronoi sites $(z_{jk})$ and cells $V_{jk} (z_{jk})$ (Eqs.~\ref{eq:25},\ref{eq:26}) as well as Voronoi cells $\upsilon_i(\theta)$ (where $i \in [1,t]$) related to all the simulation points in parameter space $\Theta = \{\theta_1, \theta_2, ..., \theta_n \} \in \mathcal{Q}$.

\item Use Eq.~(\ref{eq:21}) to calculate $\mu_{P(\theta | S=s* ) }$ related to observation point $s^*$. 

\item Use Eqs.~(\ref{eq:27} and \ref{eq:28}) to determine the maxima weighted iKernel mapping 
$\mu_{\upsilon^*}$.

\item Find a point or points corresponding to the intersection of the Voronoi cells $\upsilon^*$ (Eq.~\ref{eq:29}) in the parameter space $\mathcal{Q}$.

\end{itemize}

Note that, in principle, the intersection can be empty, i.e., $\upsilon^* = \varnothing$; thus, to find the corresponding point $\theta^*$, we use maxima weighted iKernel mapping $\mu_{\upsilon^*}$ in an additional algorithm for calculating similarity measure $l (\theta, \upsilon^*) = \psi(\theta) \mu_{\upsilon^*}$, where $\theta$ is any parameter.
The final step of the proposed algorithm can be implemented using various methods, here, we use a heuristic algorithm without a claim of best efficiency or optimality.

\subsection{Heuristic algorithm 'Tracers'}

\begin{figure}[!tpb]
\centerline{\includegraphics[width=85mm]{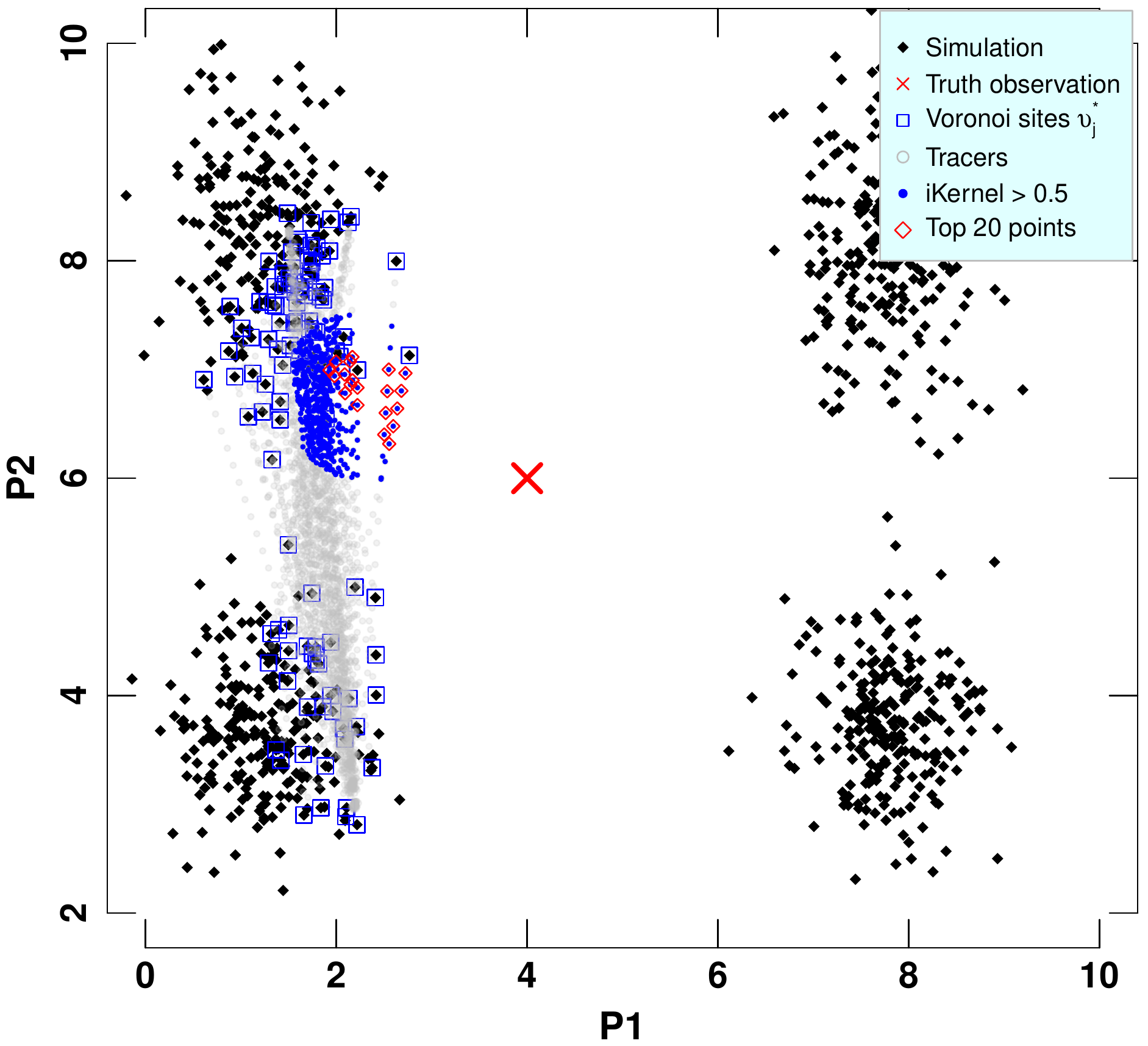}}
\caption{Finding the intersection of the Voronoi cells using tracer points (tracers) generated between Voronoi sites $\upsilon^*_j, j \in [1,t]$ from all the trees/Voronoi diagrams. The figure shows tracer points $\theta^{tr}_i$ with value of the similarity $\psi(\theta^{tr}_i) \mu_{\upsilon^*} > 0.5$, as well as the 20 points with the highest similarities.}
\label{fig:05}
\end{figure}

For reasons mentioned earlier, we seek an algorithm for determining $\theta^*$ related to $\mu_{\upsilon^*}$ that is independent of dimension of the parameter space and does not require calculation of the gradient.
For this propose, we generate points in the parameter space, transform them to a Hilbert space and, finally, measure the similarity between their iKernel mapping and the maxima weighted iKernel mapping $\mu_{\upsilon^*}$ of the observation point.
We call the generated points `tracer points' because of their similarity to the tracer bullets fired into a night sky. The algorithm consists of several steps (Algorithm~\ref{alg:Tracers}):

We first need to determine the Voronoi sites $z^*_{j} (j \in [1,t]$, Eq.~\ref{eq:27}) using Algorithm~\ref{alg:Voronoi}. Then for each site, we need to find the most distant site $z^*_{m}$ and generate points between those sites, i.e., tracer points (Fig.~\ref{fig:05}):

\begin{equation}
\{ \theta^{tr}_i \} = \{ \alpha_i \times z^*_{j} + (1 - \alpha_i) \times z^*_{m} \} \qquad \alpha_i \in [0,1] \, \forall i
\label{eq:30}
\end{equation}

Steps 2 and 3 of Algorithm~\ref{alg:Tracers} include the acquisition of the iKernel mapping $\psi(\theta^{tr}_i)$ and 
calculating the similarity between the tracer points and the maxima weighted iKernel mapping $\mu_{\upsilon^*}$ (Fig.~\ref{fig:05}):

\begin{equation}
l(\theta^{tr}_i, \upsilon^*) = \psi(\theta^{tr}_i) \mu_{\upsilon^*}
\label{eq:31}
\end{equation}


Finally, we apply linear regression to the tracer points with $\Theta_{select} = \{ \theta^{tr}_i | l(\theta^{tr}_i, \upsilon^*) > 0.5 \}$ to determine the regression coefficients for each parameter $\beta = \{ \beta_i \}$, where $i \in [1,d]$, $d$ is a dimension of the parameter space $\mathcal{Q}$. This enable us to generate points along line $\theta_{\tau} = \theta_b + \beta \times \tau$, where $\tau$ is a new variable for constructing the straight line. 
All steps are repeated until similarity (Eq.~\ref{eq:31}) ceases to change with criterion $\epsilon$.
Algorithm~\ref{alg:Tracers} gives the approximate position in the parameter space with the highest value of similarity 
$l(\theta_{\tau}, \upsilon^*) = \psi(\theta_{\tau}) \mu_{\upsilon^*}$. 


It should be noted that the heuristic algorithm Tracers is independent of the dimension of parameter space $\mathcal{Q}$ and that, rather than calculating the gradient (as in the gradient descent method), we generate tracer points, with calculation of 
$l(\theta^{tr}_i, \upsilon^*)$ for each tracer $\theta^{tr}_i$. 
It is for this reason that the algorithm is suitable for multidimensional data as well as for stochastic simulation based on branching processes.
Thus, the heuristic algorithm Tracers in combination with maxima weighted iKernel mapping based on the Voronoi diagram, gives approximate parameters for unevenly distributed sample data produced by a stochastic model.
To evaluate the efficiency of the proposed algorithms, we used multidimensional Gaussian distributed data points with a gap in the center of the distribution. 
To determine the degree to which the accuracy of an estimated parameter depends on the stochasticity of the model, we increased the noise in the simulations.

\begin{algorithm}[H]
 \KwIn{Voronoi sites $Z^* = z^*_{j} (j \in [1,t]$, Eq.~\ref{eq:27}) from the Alg.~\ref{alg:Voronoi}} 
 
 STEP 0: 
 Define initial $Z^b = \{ z^b_j \} = \{ z^*_{j} \}$  and $l_{previous} = 0$\\
 
 STEP 1: 
 \ForEach{$z^b_{j} \in Z^b$ }{  
 Find most distant site $z^b_{m} \in Z^b$ \\
 Generate tracer points: \\
 $\qquad \{ \theta^{tr}_{ij} \} = \{ \alpha_i \times z^b_{j} + (1 - \alpha_i) \times z^b_{m} \} \qquad \alpha_i \in [0,1] \, \forall i,$ \\ 
 where $i \in [1,n_{tr}]$ and $n_{tr}$ is the number of tracers for each j
} 

STEP 2: 
Run Algorithm~\ref{alg:Voronoi} again in order to calculate iKernel mapping $\psi(\theta^{tr}_{ij})$

STEP 3: 
Calculate the similarity between tracer points and maxima weighted iKernel mapping $\mu_{\upsilon^*}$ (Fig.~\ref{fig:05}): \\

$\qquad l(\theta^{tr}_{ij}, \upsilon^*) = \psi(\theta^{tr}_{ij}) \mu_{\upsilon^*}$

STEP 4: 
Extract tracers with maximal similarities: \\ 
$\qquad \theta^{top}_k, k \in [1,k_{max}]$

STEP 5: 
Using linear regression based on $\theta^{top}_k$, estimate parameter
$\theta^{est}$ with maximal similarity $l_{max}$.

STEP 6: 
\eIf {$l_{max} - l_{previous} < \epsilon$ }{ STOP }  
{ $l_{previous} = l_{max}$ \\
$Z^b = \{ z^b_k \} = \theta^{top}_k \}$
\\ Goto STEP 1 }

\KwResult{ $\theta^{est}$ and its similarity $l_{est} = \psi(\theta^{est}) \mu_{\upsilon^*}$ }

\label{alg:Tracers}

\caption{Heuristic algorithm `Tracers' to find parameter related to maxima weighted iKernel mapping $\mu_{\upsilon^*}$}

\end{algorithm}

\section{Experiments}

\subsection{Synthetic data}

\begin{figure}[!tpb]
\begin{center}
\centerline{\includegraphics[width=77mm]{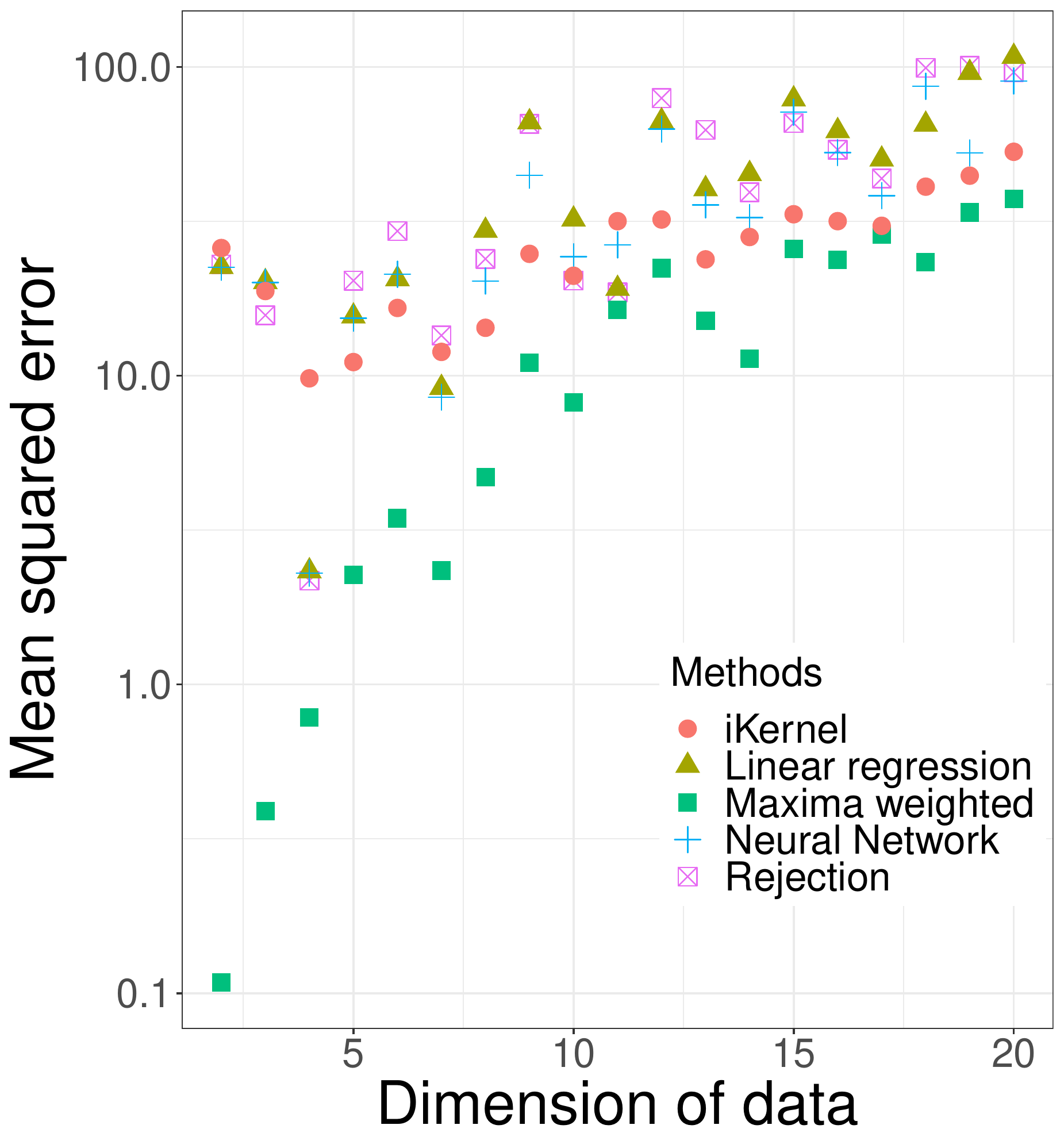}}
\caption{Mean squared error vs dimension of data  with stochastic term $\eta_{stoch} = 0.6$ from ABC estimations using different methods: 
linear regression, maxima weighted iKernel mapping, Isolation Kernel ABC, neural network and rejection ABC.}
\end{center}
\label{fig:07}
\end{figure}

To check the efficiency of the proposed algorithm we used multidimensional data from two models:
the first is a Gaussian function with unevenly distributed points and a gap at the center of the distribution (region of interest);
the second is a linear model with stochastic term $\eta_{stoch}$ (see Application 2 for details).
In this research, we eschew the generation of new points and compare methods based on a given sample of points using the mean squared error (MSE). 
This is simply to avoid the effect of the sampling method and to compare ABC methods based solely on a given sample, using only four methods in the comparison: rejection ABC \cite{ABC_rejection} (Rejection in Fig.~\ref{fig:07}), and ABC using the regression-based correction methods that employ either local linear regression \cite{ABC_lin_reg} (Linear regression in Fig.~\ref{fig:07}) or neural networks (NN)  \cite{ABC_NN} (Neural Network in Fig.~\ref{fig:07}). For our calculations, we used the `abc' package in R that includes all of the above methods. The proposed method and kernel ABC based on Isolation Kernel are shown denoted in Fig.~\ref{fig:07} as Maxima weighted and iKernel, respectively.

The results of our synthetic experiments are presented in the Application 2. For the non-linear model using Gaussian functions for each dimension iKernel ABC, and ABC based on Maxima weighted iKernel mapping gave the best results for all simulations with different dimensions (Maxima weighted iKernel mapping typically had 3.5-20 times the accuracy of iKernel ABC (see Fig.~\ref{fig:app_02_2})). 
Results for the linear model depended on the stochastic term: 
for $\eta_{stoch} = 0$ linear regression gave the exact values with MSE~$<10^{-20}$;
for $\eta_{stoch} = 0.3$ and range of output $Y \in [0,10]$ ABC based on linear regression, NN and Maxima weighted iKernel mapping yielded results with similar MSE, while iKernel and rejection ABC produced MSEs that were several times larger;
for $\eta_{stoch} \ge 0.6$ (and the same output range) MSE for Maxima weighted iKernel mapping was best in all cases (Fig.~\ref{fig:07}).

\subsection{Cancer cell evolution}

The cancer cell evolution model serves as an interesting example of branching processes with a high probability of cell population extinction. 
In this research the tugHall simulator from \cite{tugHall_bioinf2020} was used, together with the simulation dataset from \cite{Dataset_tugHall2021}. 
The model has 27 parameters, seven of which were fixed based on expert estimation \cite{tugHall_bioinf2020}. 
The remaining 20 parameters represent the hallmark--gene relationship that is an analogue to gene--phenotype relations.  
The dataset includes the results of 9,600,000 simulations using different models, initial conditions, and input parameters. Here, the model with threshold metastatic transformation was used with the initial mutated cell in the pull of 1000 normal cells. 
This sample has the largest number of successful simulations (34,602) and among them the largest number non-zero outputs (34,059) from 400,000 simulations. 

Based on the given sample of the dataset and the VAF of a patient from the Cancer Genome Atlas (TCGA) database \cite{TCGA} as an observation (ID record TCGA-AF-5654-01A-01D-1657-10), parameter estimation was performed using the various methods. For rejection ABC, NN ABC and linear regression ABC, the mean, median and mode values were used in 100 repetitions of the simulation for each parameter set.
Unfortunately, only one measure of the central tendency produced a non-zero output for each method 
(denoted in parentheses in Table~\ref{table:ABC}) 
while the others led to extinction for all repetitions. 
It is for this reason that Table~\ref{table:ABC} shows the results of simulations for only one successful attempt at determining the parameter set, excluding iKernel ABC which led to extinction in all 100 simulations. 

The observation data are represented as a vector with four VAF values for each gene (APC, KRAS, TP53, and PIK3CA) related to colorectal cancer. 
Corresponding distributions of the simulation output for each gene in the form of box plots were produced in Application 3. Table~\ref{table:ABC} shows the statistical characteristics of the results of the simulations for each method. In addition to MSE, the portion of successful simulations with survival cells is represented ($\eta_{succeed}$ ). Standard deviation $\sigma$ shows the degree to which the simulation data are concentrated around their center (for NN ABC $\sigma = 0$ because of the low probability of successful simulation).
The energy distance, E-dist, which is the statistical distance between observation and simulation, is also included in Table~\ref{table:ABC} and was calculated using the `energy' R package.
\cite{E_dist2013,E_dist_2017}.

\begin{table}[t]
\begin{center}
\caption{Simulation statistics based on parameter estimation using the different methods: $\eta_{succeed}$ is the portion of successful simulations with non-zero output, $\sigma$ is the standard deviation, and E-dist is the energy distance.}
\label{table:ABC}
\begin{tabular}{lcccr}
\hline 
Method       & MSE &    $\eta_{succeed}, \% $     & $\sigma$       & E-dist \\
\midrule
Rejection \\ (Median)    & 6956  &    38~\%                              & 212  & 1.616 \\
Neural \\ Network \\ (Mean) & 6999  &    7~\%                              & 0  & 1.464 \\
Linear \\ Regression \\ (Mode) & 1975  &    43~\%                              & 206  & 0.847 \\
iKernel &    $\times$  &    $\times$                          & $\times$  & $\times$ \\
Maxima\\ weighted   & 68  &    52~\%                              & 97  & 0.086 \\
\hline  
\end{tabular}
\vskip -2in
\end{center}
\end{table}

\begin{figure}[h]
\begin{center}
\centerline{\includegraphics[width=72mm]{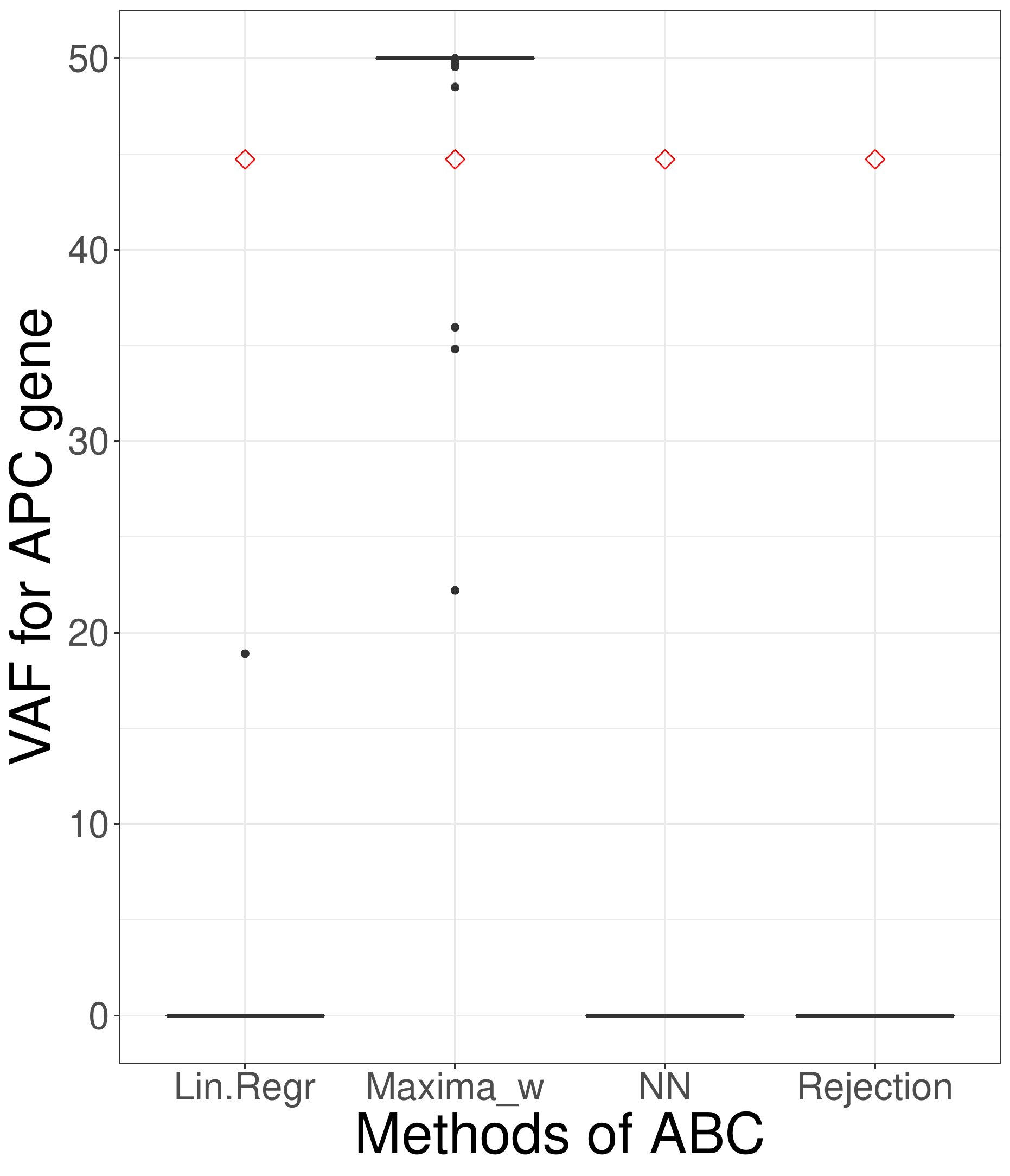}}
\caption{Box plot of VAF of APC gene obtained by various methods. The observation point is shown as a diamond.}
\label{fig:app_03_APC}
\end{center}
\end{figure}

\begin{figure}[h]
\begin{center}
\centerline{\includegraphics[width=72mm]{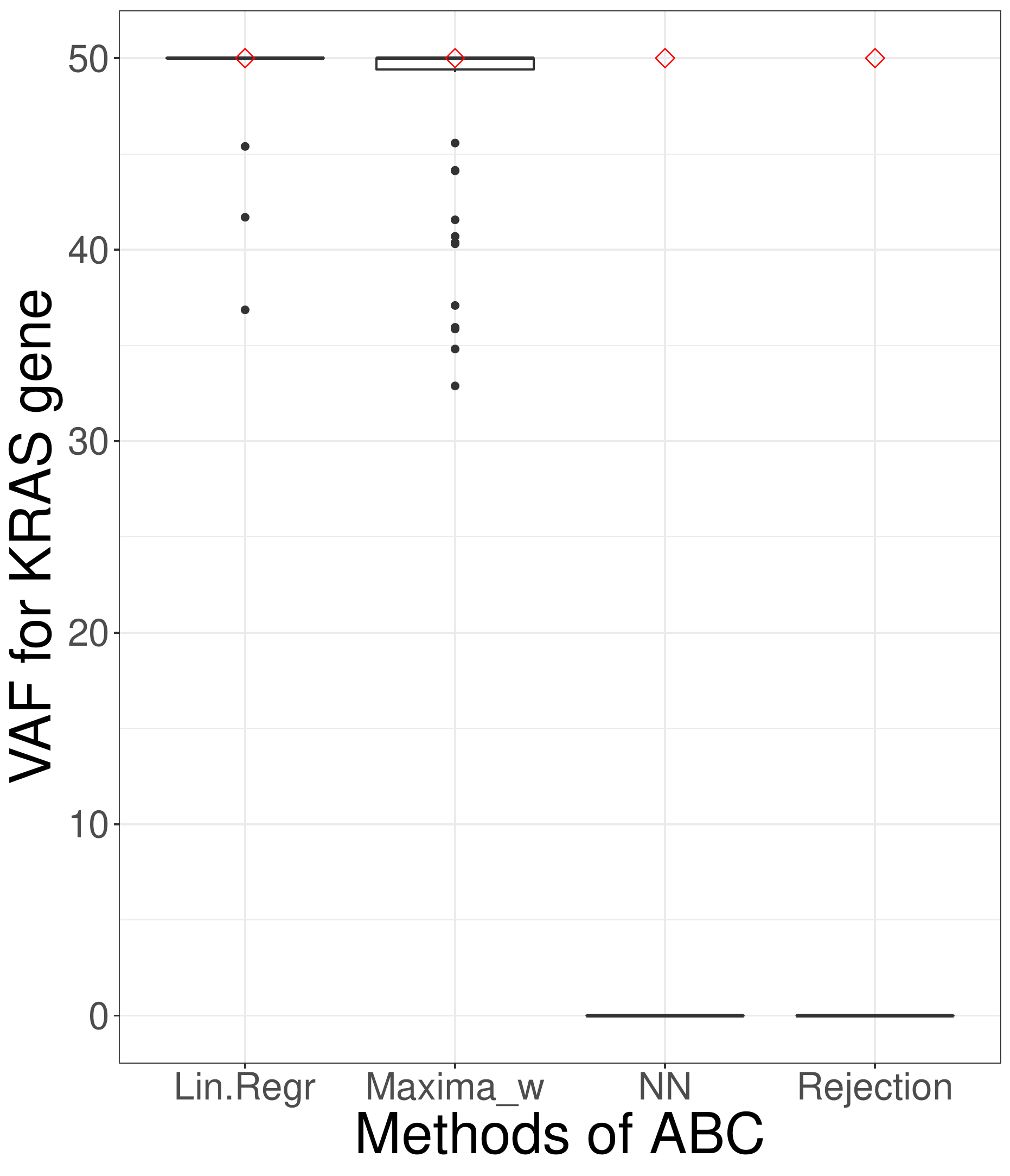}}
\caption{Box plot of VAF of KRAS gene obtained by various methods. The observation point is shown as a diamond.}
\label{fig:app_03_KRAS}
\end{center}
\end{figure}

\begin{figure}[h]
\vskip 0.55in
\begin{center}
\centerline{\includegraphics[width=72mm]{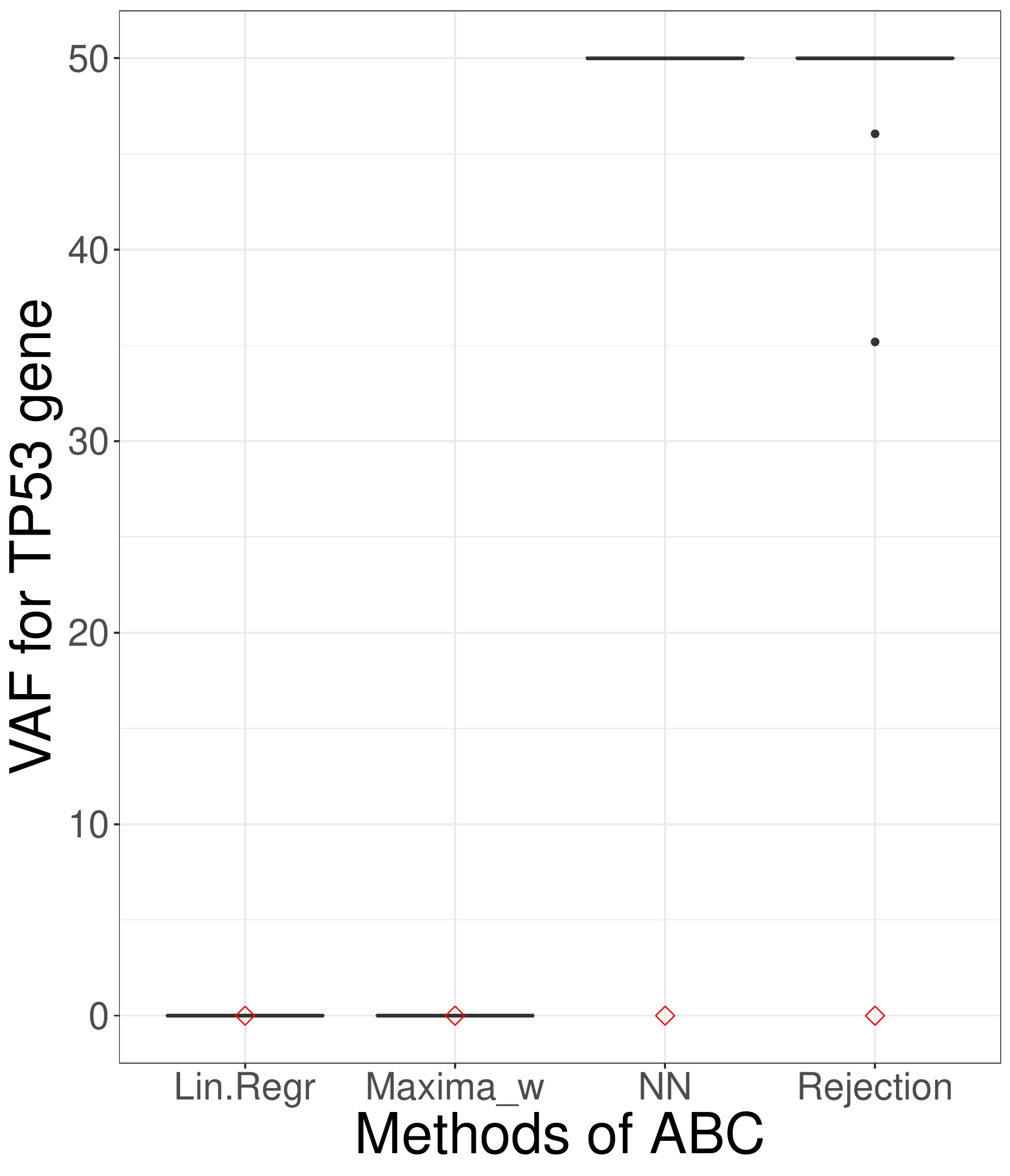}}
\caption{Box plot of VAF of TP53 gene obtained by various methods. The observation point is shown as a diamond.}
\label{fig:app_03_TP53}
\end{center}
\end{figure}

\begin{figure}[h]
\vskip -0.2in
\begin{center}
\centerline{\includegraphics[width=72mm]{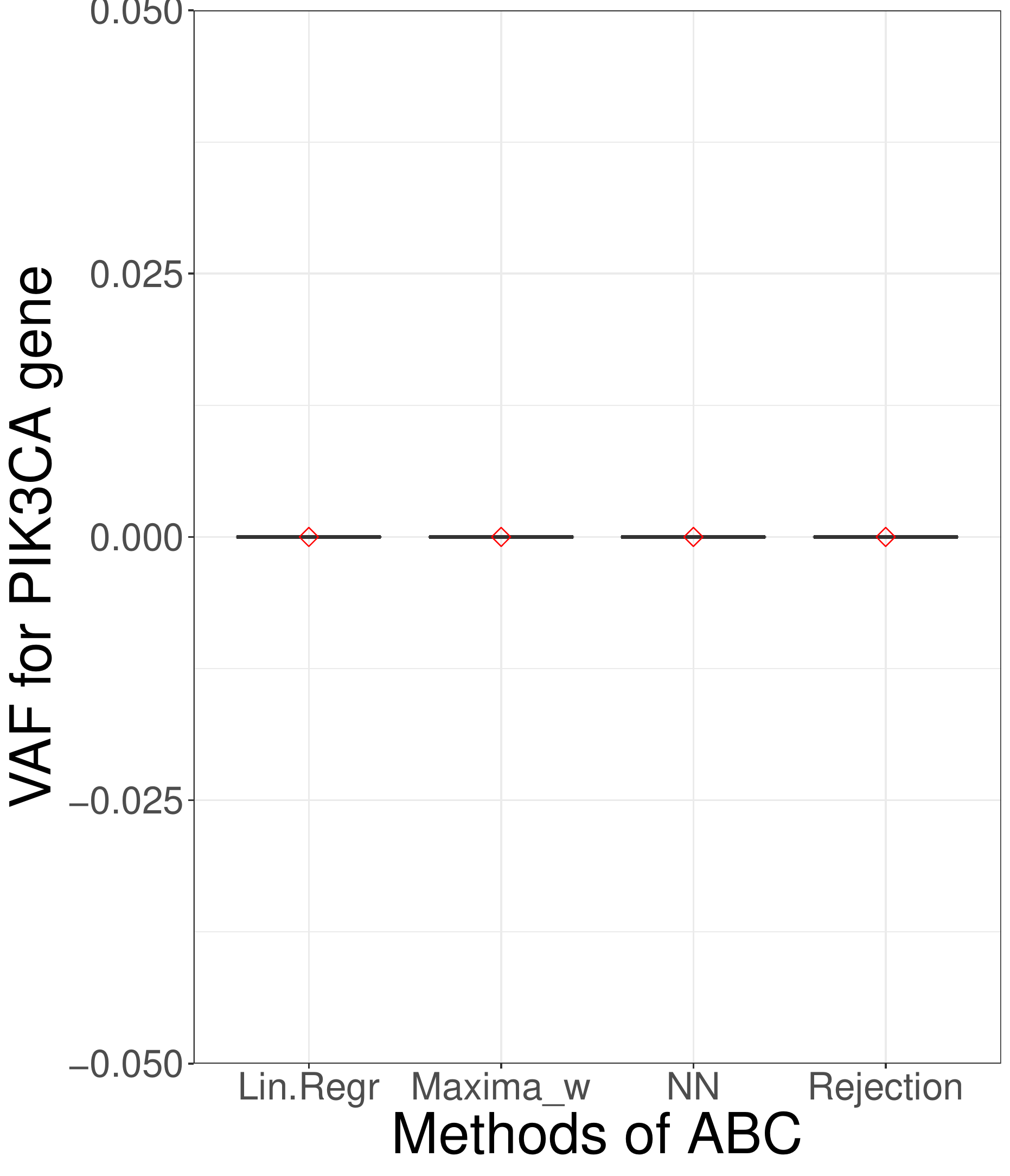}}
\caption{Box plot of VAF of PIK3CA gene obtained by various methods. The observation point is shown as a diamond.}
\label{fig:app_03_PIK3CA}
\end{center}
\end{figure}

According to Table~\ref{table:ABC}, the ABC method based on Maxima weighted iKernel mapping shows the best accuracy. 
Rejection ABC and ABC based on correction with NN have a large mean squared error and a high energy distance, as 
they are in good agreement only with the VAF of the PIK3CA gene (Fig.~\ref{fig:app_03_PIK3CA} in Application 3). 
ABC based on correction with linear regression diverges from the observation for the VAF of the APC gene (Fig.~\ref{fig:app_03_APC}), which is the most frequent driver gene in colorectal cancer evolution \cite{pan_cancer2020}.
Only the proposed method gives good agreement for the VAF of all four genes (Figs.~\ref{fig:app_03_APC},\ref{fig:app_03_KRAS},\ref{fig:app_03_TP53}, and \ref{fig:app_03_PIK3CA})

\section{Conclusion}

We have proposed a heuristic ABC based on the part of Isolation Kernel mean embedding with 
maximal weights $\mu_{\upsilon^*}$ (Eqs.~\ref{eq:25},\ref{eq:26},\ref{eq:27}, and \ref{eq:28}).
Isolation Kernel is implemented with a Voronoi diagram algorithm, which makes possible
the explicit transformation of simulation row data to a Hilbert space corresponding to the model's parameters. 
In the simplest case, the interpretation of $\mu_{\upsilon^*}$ is the intersection of all the Voronoi cells of $\mu_{\upsilon^*}$ corresponding to an observation point (Eq.~\ref{eq:29}).

The proposed method allows parameter estimation with good accuracy 
even when the observation point lies in the sparse region of a given data sample. 
Maxima weighted iKernel mapping is also effective in dealing with the stochasticity of the simulation data, as well as high dimensionality. 

Application of the method to synthetic 2--20 dimension data with a stochastic term and the tugHall simulator of cancer cell evolution showed much better results in comparison to well-known ABC methods.
Combining Maxima weighted iKernel mapping with sampling techniques is one of the possible ways to develop and extend the method.

\section*{Acknowledgements}
The author wishes to thank Prof. Takashi Washio for his detailed discussion of the theoretical part of this research, as well as for all the valuable critical comments.
\newpage

\bibliographystyle{natbib}
\bibliography{./References.bib}

\section*{Application 1. Isolation Forest}

The Isolation Forest (iForest) algorithm was initially proposed in \cite{AnomalyDetection2012,IsolationForest2008}. 
The authors defined anomalous data points as points in a sparse region of dataset $D \in R^d$ as compared to normal points that appear in dense regions.
Since anomalies are located in sparse areas, they are easier to `isolate'. 
Isolation Forest builds an ensemble of Isolation Trees ($iTrees$) based on multiple subsets $\mathcal{D}_i \subset D$, and anomalies are  all the $x \in D$ points in the full dataset $D$ that have shorter average path lengths $\hat{h}(x)$ on the $iTrees$.

In \cite{AnomalyDetection2012}, the authors use a set of experiments to prove that iForest offers the following advantages:

\noindent
- low linear time complexity and a small memory requirement due to the small size $\xi = |\mathcal{D}|$ of subsets $\mathcal{D}_i, i \in [1..t]$ and the finite number of $iTrees$ $t$;

\noindent
- the capacity to deal with high-dimensional data $\mathcal{D} \in R^d$, where $d$ is dimensionality;

\noindent
- the ability to be trained with or without anomalies in the $\textbf{Training Set =} \bigcup_{i=1}^t \mathcal{D}_i$;

\noindent
- the ability to provide detection results with different levels of granularity without re-training at the \textbf{Evaluation stage}.

In order to isolate a data point the iForest algorithm (Algorithm~\ref{alg:iForest}) recursively generates partitions $\mathcal{H}_i$ of sample $\mathcal{D}_i$ by randomly selecting an attribute $q \in [1..d]$ and then randomly selecting a split value $p$ for the attribute that is between the minimum $q_{min}$ and maximum $q_{max}$ values allowed for that attribute.

$iTree_k$ at \textbf{training step} $k$ has information on recursive partitioning $\mathcal{H}_k$.
The length of path $h_k(x)$ in the $iTree_k$ is defined as the number of partitions required to isolate point $x$ within tree $iTree_k$ to reach a terminating leaf-node, starting from the root-node.
Let ${D} = \{ x_1, ..., x_n \} \in R^d$ be a set of $d$-dimensional points, $\mathcal{D} = \{ z_1, z_2, .., z_{\xi} \}$, and $\mathcal{D} \subset D$ be a subset of ${D}$ where the size of subset $|\mathcal{D}| = \xi \ll n$. 
Thus, $iTree$ is defined as a data structure with the following features:

\noindent
- for each node $V$ in the $iTree$, $V$ is either a leaf-node with no child or a branching-node with one partition and exactly two daughter nodes ($V_l$, $V_r$: the left and right branches, respectively).

\noindent
- a partition at node $V$ consists of an attribute $q$ and a splitting value $p$ such that the condition $x^q < p$ determines the traversal of a data point $x$ to left branch $V_l$, and otherwise to right branch $V_r$.

\vspace*{12pt}

\begin{algorithm}[H]
 \KwData{Multidimensional Dataset $D = \{x_1, x_2, ..., x_n \} \in R^d$ }
 \KwIn{ $t$ - number of trees, $\xi$ - subsampling size, $\xi \ll n$ } 
{\bf I) Training Stage} \\ 
  \For{k = 1 .. t}{
	{\bf Make Isolation Tree} $iTree_k$:\\
	Randomly produce subset $\mathcal{D} \subset D$, where $\mathcal{D} = \{z_1, z_2, ..., z_j, ..., z_{\xi} \}$ \\
	Randomly split space $R^d$ by partitions $\mathcal{H}_k$ \\
	until all $z_j \in \mathcal{D}$ are isolated [based on $iTree$ algorithm] \\
	$iTree_k =$ a tree of partitions $\mathcal{H}_k$\\
}
\KwOut{ Isolation Forest = a set of $iTree_k$, where $k \in [1..t]$} 

{\bf II) Evaluation Stage} \\ 
\ForEach{$x \in D$ }{
	\For{k = 1 .. t}{
		find the corresponding leaf-node for $x$ in $iTree_k$ \\
		calculate $h_k(x)$ - length of path in $iTree_k$ \\
	}
	find the average $\hat{h}(x) = \frac{1}{t} \sum^{t}_{k=1} h_k(x)$
}

\KwResult{ $\hat{h}(x)$ - average  path length in the iForest} 
\caption{Isolation Forest algorithm to detect anomalies}
\label{alg:iForest}
\end{algorithm}

\vspace*{4pt}

In order to construct an $iTree$, the algorithm recursively divides subset $\mathcal{D}$ by randomly selecting an attribute $q$ and a split value $p$, until either (i) the node has only one point or (ii) all points at the node have the same coordinates.
Once each point in subset $\mathcal{D}$ is isolated at one of the leaf-nodes, $iTree$ is fully grown and the \textbf{training stage} is finished.

Algorithm~\ref{alg:iForest} shows the two stages of Isolation Forest - the training stage and the evaluation stage.
It is clear that the anomalous points are those points with a shorter path length in the tree, where the path length $h(x_i)$ of point $x_i \in D$ is defined as the number of edges $x_i$ traversed from the root vertex to reach a leaf-node.
Algorithm~\ref{alg:iForest} shows that the training stage is needed to construct the $t$ $iTree$s for $t$ subsets $\mathcal{D}_k, k \in [1..t]$. The set of $iTree$s is the Isolation Forest.
During the evaluation stage, iForest measures the path length in each $iTree$ $h_k(x), k=[1..t]$ for each point $x \in D$.

\section*{Application 2. Results of synthetic simulation}

For our synthetic simulations, we used multi-dimensional data from two models, one based on a linear function, the other based on a Gaussian function.
Each model has dimensionality $d \in [2,20]$, input parameter $X =  \{ x_1, ..., x_d  \}$ and output data with the same dimensionality
$Y = \{ y_1, ..., y_d  \}$. 
To simplify consideration of the results, we defined each element of the output vector as a function of only one input variable $y_i = f(x_i)$, where  $i \in [1..d]$. 
For each model, the values of parameters $x_{01}, ..., x_{0d}$ are to be determined. Thus, the linear model is defined as
\begin{equation}
Y_{lin}(X) 
= \left \{ 
\begin{array}{l}
\alpha_1 \cdot (x_1-x_{01}) + \eta_{1,stoch}, \\
..., \\
\alpha_i \cdot (x_i -x_{0i})  + \eta_{i,stoch}, \\ 
..., \\
\alpha_d \cdot (x_d -x_{0d})  + \eta_{d,stoch}  
\end{array}
\right \}
\label{eq:linmod}
\end{equation}
where $\eta_{i,stoch}$ is a stochastic term of $i^{th}$ dimension. 

The Gaussian-based model is:
\begin{equation}
Y_{Gauss}(X) 
= \left \{ 
\begin{array}{l}
e^{ - \alpha_1 \cdot (x_1 - x_{01})^2}, \\
..., \\
e^{ - \alpha_i \cdot (x_i - x_{0i})^2}, \\ 
..., \\
e^{ - \alpha_d \cdot (x_d - x_{0d})^2} 
\end{array}
\right \}
\label{eq:expmod}
\end{equation}

\begin{figure}[h]
\begin{center}
\centerline{\includegraphics[width=85mm]{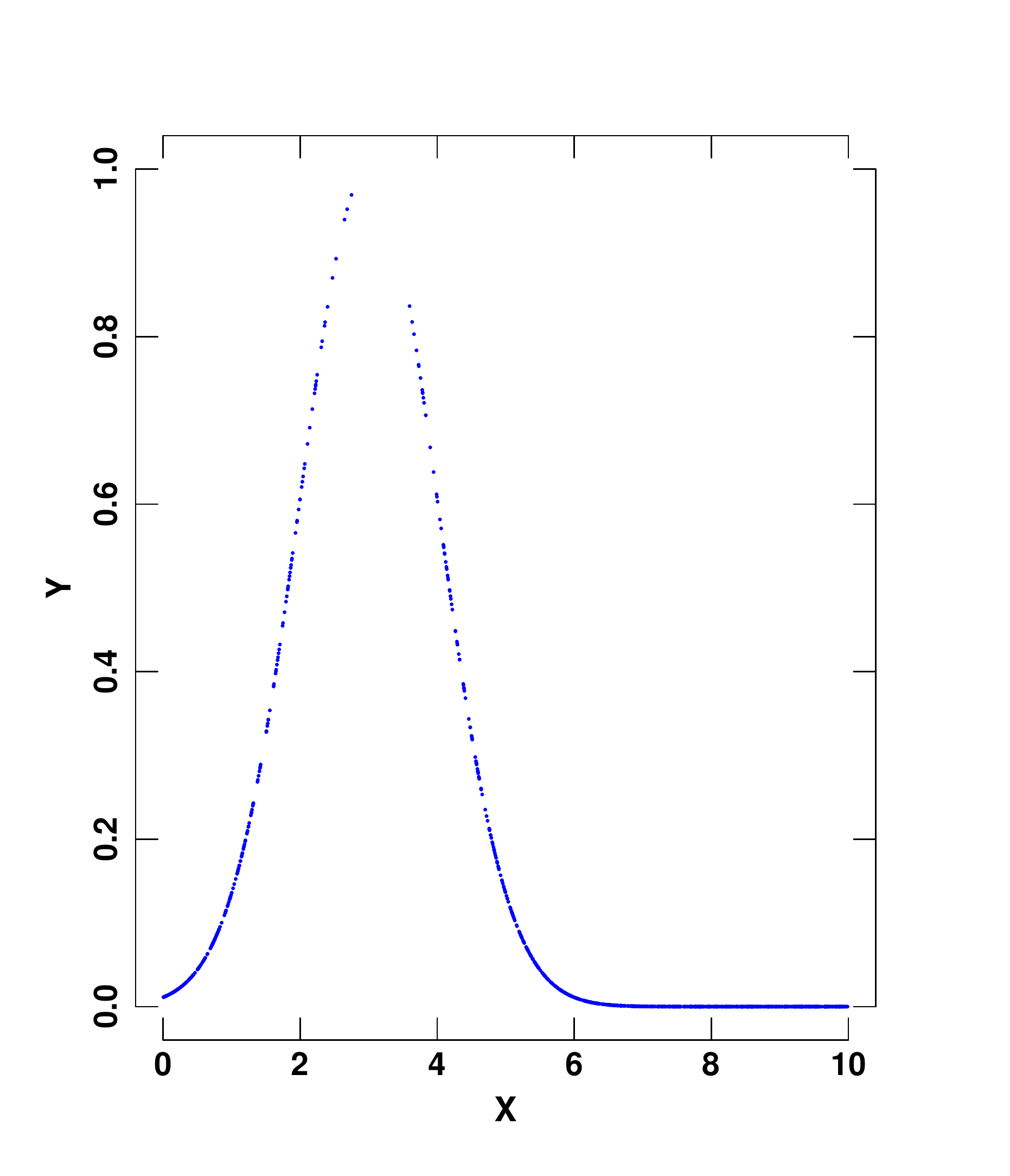}}
\caption{Row data for the model based on the Gaussian function and lower probability of generation at the maximum (point of interest).}
\label{fig:app_02_1}
\end{center}
\end{figure}

The number of points in the sample depends on the distance to point of interest $x_{0i}$ (see Fig.~\ref{fig:app_02_1}, for example)
in order to establish sparse regions and imitate the lack of information around the observation point.

\begin{figure}[h]
\begin{center}
\centerline{\includegraphics[width=85mm]{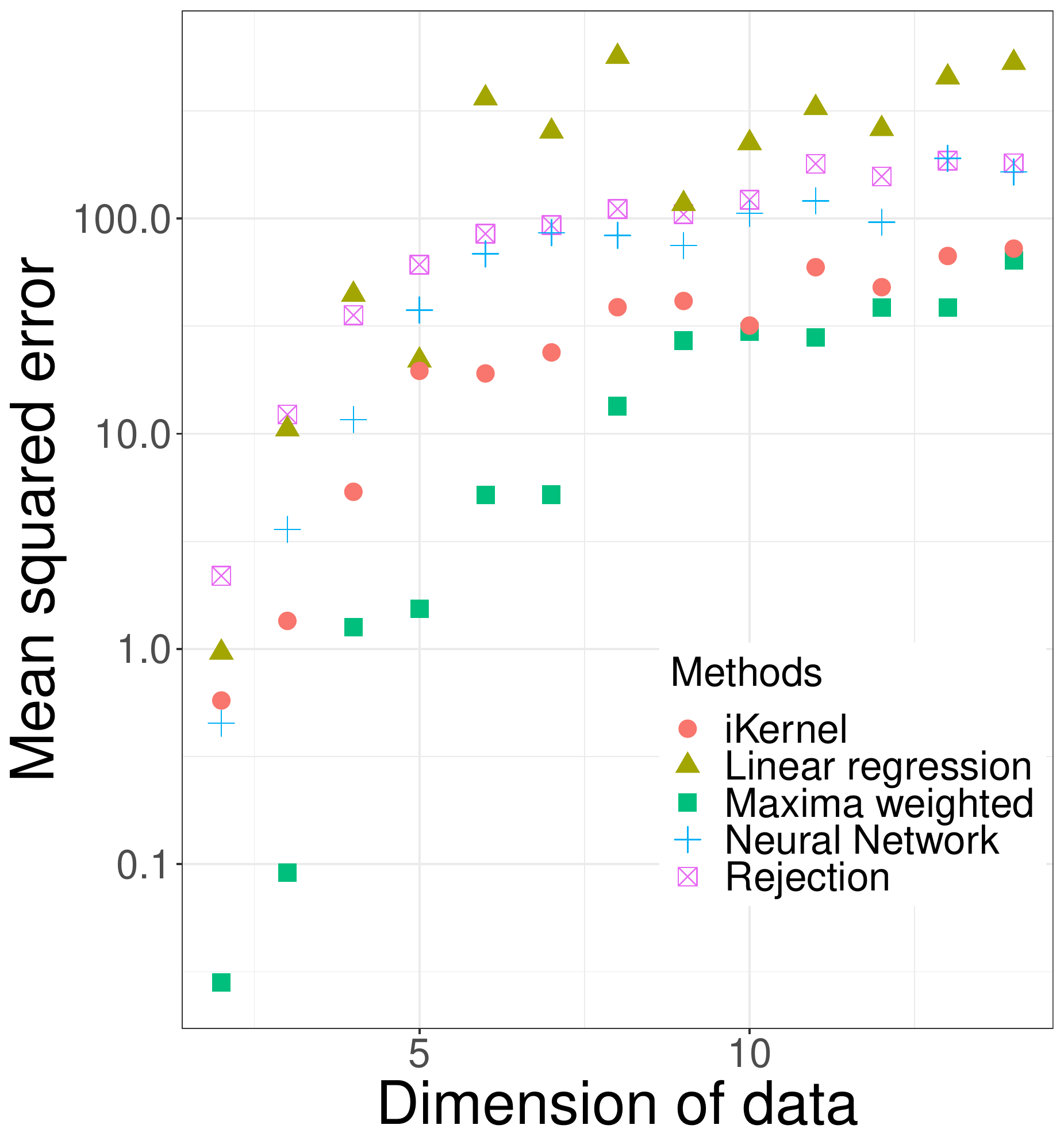}}
\caption{Mean squared error of ABC estimates vs dimensionality of data for the Gaussian-based model.}
\label{fig:app_02_2}
\end{center}
\end{figure}

\begin{figure}[h]
\begin{center}
\centerline{\includegraphics[width=85mm]{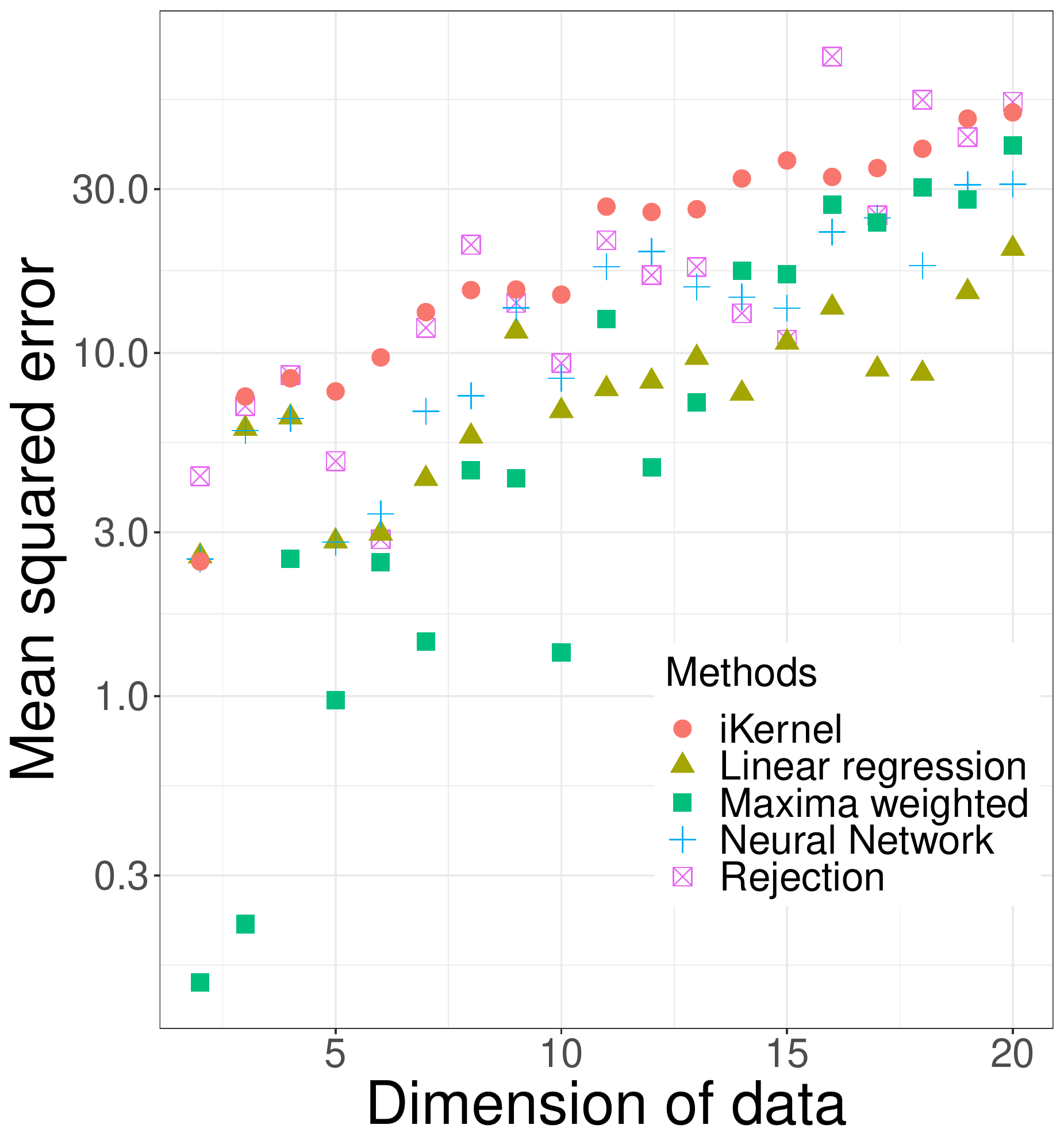}}
\caption{Mean squared error of ABC estimates vs dimensionality of data for the model using linear function with stochastic term $\eta_{stoch} = 0.3$.}
\label{fig:app_02_3}
\end{center}
\end{figure}

We used four methods for parameter estimation: rejection ABC \cite{ABC_rejection} (Rejection in Figs.~\ref{fig:app_02_2} and \ref{fig:app_02_3}), ABC using regression-based correction methods that employ either local linear regression \cite{ABC_lin_reg} (Linear regression in Figs.~\ref{fig:app_02_2} and \ref{fig:app_02_3}) or neural networks \cite{ABC_NN} (Neural Network in Figs.~\ref{fig:app_02_2} and \ref{fig:app_02_3}). 
The `abc' package in R with default values was used for calculations. 
We also attempted to adjust the internal parameters to improve the results although this is redundant for such toy tasks.
The proposed method and kernel ABC based on Isolation Kernel are shown in Figs.~\ref{fig:app_02_2} and \ref{fig:app_02_3}) as Maxima weighted and iKernel respectively. 

For the non-linear model using Gaussian functions, iKernel ABC and ABC based on Maxima weighted iKernel mapping produced the best results for all simulations and dimensions (the accuracy of Maxima weighted iKernel mapping was typically $3.5-20$ times that of iKernel ABC (see Fig.~\ref{fig:app_02_2}) with the exception of a few points).
No surprise, the worst MSE values were for ABC based on linear regression since the model is significantly non-linear. 

The results for the linear model depended on the stochastic term used the following: 
for $\eta_{stoch} = 0$, linear regression gives exact values with MSE~$<10^{-20}$;
for $\eta_{stoch} = 0.3$ and range of output $Y \in [0,10]$, ABC based on linear regression, NN and Maxima weighted iKernel mapping yield results with a similar MSE(Fig.~\ref{fig:app_02_3}), while iKernel and rejection ABC produce an MSE that is several times larger;
for $\eta_{stoch} \ge 0.6$ (and the same output range), the MSE for Maxima weighted iKernel mapping is best in all cases (Fig.~\ref{fig:07}).  
Unfortunately, iKernel ABC showed an accuracy similar to that of the rejection algorithm, which is why one might expect the same behavior in the cancer simulation case.

\section*{Application 3. Box plots of the results of simulations of cancer cell evolution}

\end{quote}

\end{document}